%% file: collas2026_conference.tex
\documentclass{article} % For LaTeX2e
\usepackage{collas2026_conference,times}
\usepackage{easyReview}
\usepackage{booktabs}
\usepackage{multirow}
\usepackage{amssymb}

\usepackage{caption}
\usepackage{pgfplots}
\pgfplotsset{compat=1.18}

\usepackage[dvipsnames]{xcolor}
\usepackage[normalem]{ulem}

\input{math_commands.tex}

\usepackage{hyperref}
\hypersetup{
    colorlinks=true,
    linkcolor=red,
    filecolor=magenta,
    urlcolor=blue,
    citecolor=purple,
    pdftitle={Continual Distillation Learning for Rehearsal-Free Class-Incremental Learning via Decoupled Prompting},
    pdfpagemode=FullScreen,
    pdfauthor={Qifan Zhang, Yunhui Guo, Yu Xiang},
    }

\title{Continual Distillation Learning for Rehearsal-Free Class-Incremental Learning via Decoupled Prompting}

\author{Qifan Zhang, Yunhui Guo, Yu Xiang  \\
Intelligent Robotics and Vision Lab\\
The University of Texas at Dallas\\
\texttt{\{qifan.zhang,yunhui.guo,yu.xiang\}@utdallas.edu}
}

 \preprintcopy % Uncomment for the preprint version, but NOT for submission.

\begin{document}

\maketitle

\begin{abstract}
Prompt-based continual learning has shown strong performance in rehearsal-free class-incremental learning by adapting learnable prompts while freezing a pre-trained Vision Transformer (ViT) backbone. However, the effect of backbone scale remains underexplored. We observe that larger ViT backbones consistently yield better continual learning performance, which motivates us to study how to transfer such capability from a larger model to a smaller one. In this paper, we introduce Continual Distillation Learning (CDL), a new setting for knowledge distillation in rehearsal-free prompt-based continual learning. We show that conventional distillation methods provide only limited gains in CDL, mainly because task-specific prompts are forced to encode both continual adaptation and distillation knowledge, while lacking a persistent mechanism for cross-task knowledge transfer. To address this problem, we propose Decoupled Continual Distillation Learning (D-CDL), which introduces persistent Knowledge-Distillation prompts (KD-prompts) and a dedicated KD branch to explicitly decouple distillation from task adaptation. The proposed KD-prompts are propagated across tasks as a global carrier of teacher knowledge, while the original prompts remain responsible for continual learning. D-CDL is simple, general, and can be integrated into various prompt-based continual learning frameworks. Extensive experiments on Split CIFAR-100 and Split ImageNet-R across four representative continual learning methods show that D-CDL consistently outperforms existing distillation baselines and substantially improves student performance under different teacher-student settings.~\footnote{Project website: \url{https://irvlutd.github.io/CDL/}}
\end{abstract}

\section{Introduction}

Continual Learning (CL) aims to enable models to learn a sequence of tasks continuously while retaining knowledge acquired from previous tasks. In class-incremental continual learning~\cite{rebuffi2017icarl}, for example, data from new classes arrive sequentially, and the model must learn these classes one task after another during training, while being evaluated on all previously seen classes at test time. Therefore, a desirable CL model should not only acquire new knowledge effectively but also mitigate catastrophic forgetting of prior tasks.

Recently, with the rapid development of large-scale vision models, prompt-based continual learning methods have achieved remarkable progress in rehearsal-free CL, including representative approaches such as Learning to Prompt (L2P)~\citep{wang2022learning}, DualPrompt~\citep{wang2022dualprompt}, and CODA-Prompt~\citep{Smith2023CODAprompt}. These methods typically leverage a pre-trained Vision Transformer (ViT)~\citep{ViT2020image} as a frozen backbone and shift continual adaptation to learnable prompts, which are usually selected from a prompt pool through specific retrieval or selection mechanisms. By preserving the pre-trained backbone and only optimizing prompts, prompt-based CL methods have achieved state-of-the-art performance in class-incremental learning.

\begin{figure}
  \centering
  \includegraphics[width=1\linewidth]{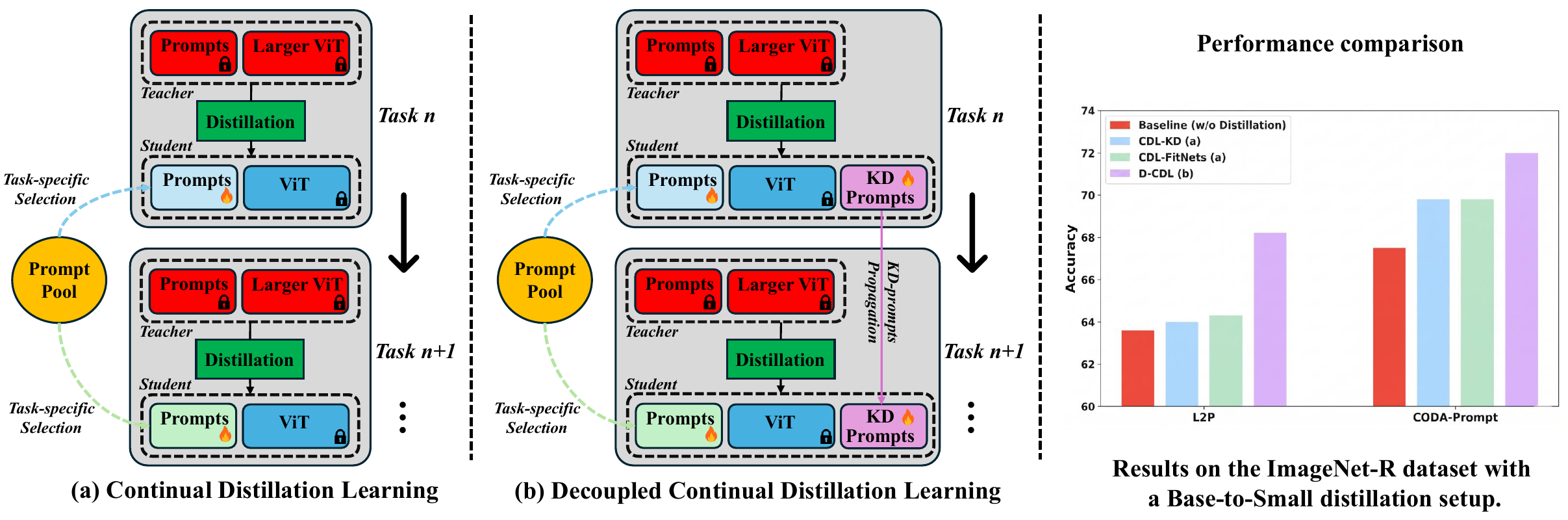}
  \vspace{-15pt}
  \caption{Illustration of (a) Continual Distillation Learning (CDL) and (b) our Decoupled Continual Distillation Learning (D-CDL). 
In CDL, standard prompt-based continual learning directly uses selected prompts to absorb both task-specific knowledge and teacher knowledge, which leads to objective entanglement and weak cross-task knowledge transfer. 
In contrast, D-CDL introduces persistent KD-prompts to decouple distillation from task adaptation, enabling globally continuous teacher-to-student knowledge transfer across tasks. 
Right: performance comparison on the ImageNet-R dataset under a ViT-Base-to-ViT-Small distillation setting.
}
\vspace{-4pt}
  \label{fig:intro}
  \vspace{-4mm}
\end{figure}

However, existing prompt-based CL methods mainly focus on prompt design while fixing the ViT backbone, most commonly using a ViT-Base model. In contrast, the effect of backbone scale itself has received much less attention. We observe that the backbone plays a critical role in continual learning performance: a larger ViT backbone often yields significantly better CL results, as shown in Appendix~\ref{appendix:ViT_size}. This observation motivates us to study a new and practically important problem: \emph{how to transfer the continual learning capability of a larger prompt-based CL model to a smaller one}. To this end, we propose a new research problem, termed \textbf{Continual Distillation Learning (CDL)}, which studies knowledge distillation from a larger teacher to a smaller student in rehearsal-free prompt-based continual learning. As illustrated in Fig.~\ref{fig:intro}(a), under the CDL setting, a larger and better-performing teacher model is first trained on the current task, and then frozen to guide the training of a smaller student model on the same task. This process is repeated throughout the task-incremental learning sequence, where both teacher and student can access only the data from the current task. Such a setting is practically attractive because it allows us to continuously distill knowledge from a stronger large model into a smaller one, thereby improving efficiency during inference while preserving continual learning ability. Unlike standard knowledge distillation in a static setting, CDL must continuously transfer knowledge from a stronger teacher to a smaller student under rehearsal-free class-incremental learning, where only current-task data are available and the student must preserve previously acquired knowledge. This makes CDL a distinct and practically important problem that jointly involves continual adaptation, knowledge transfer, and efficiency-oriented model compression.

A natural question is whether existing knowledge distillation methods, such as logit distillation method KD~\citep{hinton2015distilling} and feature distillation method FitNets~\citep{romero2014fitnets}, can be directly applied to the CDL problem as CDL-KD and CDL-FitNets. 
Unfortunately, our experiments show that their performance gains are limited, as illustrated by the performance comparison in Fig.~\ref{fig:intro}. 
We argue that this difficulty mainly arises from two fundamental issues in prompt-based CDL. First, in prompt-based continual learning, the ViT backbone is frozen and only the selected prompts are optimized. As a result, when standard distillation objectives are introduced, the same prompts are forced to encode both task-specific classification knowledge and teacher-to-student distillation knowledge. This entangles two inherently different objectives within a single prompt mechanism, making optimization less effective. Second, prompt-based CL methods typically construct a new set of prompts for each incoming task through task-specific prompt selection. While this design is effective for continual learning, it is not well suited for distillation. In particular, the teacher knowledge absorbed in the current task may not be reliably preserved in future tasks, because the prompts involved in one task may not be reused in subsequent tasks. In other words, prompt selection provides task specialization for continual learning, but lacks a persistent carrier for cross-task distillation knowledge. Unlike task-specific classification knowledge, however, teacher knowledge should be accumulated progressively across tasks and preserved in a globally consistent manner.

To address these challenges, we propose \textbf{Decoupled Continual Distillation Learning (D-CDL)}, a simple yet effective framework that decouples continual learning prompts from distillation-specific prompts. Specifically, as shown in Fig.~\ref{fig:intro}(b), we introduce a new set of Knowledge-Distillation prompts (KD-prompts) to separately model teacher knowledge, while preserving the original prompts for task-specific continual adaptation. Different from standard task-specific prompts, the proposed KD-prompts are continuously propagated across tasks, serving as a global and persistent carrier of distillation knowledge throughout the entire continual learning process. Importantly, D-CDL is a general framework that can be readily integrated into a wide range of existing prompt-based CL methods. Experimental results in Fig.~\ref{fig:intro} show that D-CDL consistently outperforms standard KD and FitNets by a clear margin, demonstrating the effectiveness of decoupling continual learning and distillation in prompt-based rehearsal-free continual learning.

% In summary, our main contributions are as follows:
% \begin{itemize}
%     \item We introduce \textbf{Continual Distillation Learning (CDL)}, a new problem setting that studies knowledge distillation from large to small models in rehearsal-free prompt-based continual learning.
%     \item We analyze why existing distillation methods are suboptimal in prompt-based CDL, and identify two key challenges: objective entanglement in prompts and the lack of cross-task continuity for distillation knowledge.
%     \item We propose \textbf{Decoupled Continual Distillation Learning (D-CDL)}, which introduces persistent KD-prompts to decouple task adaptation from distillation and enables global knowledge transfer across tasks.
%     \item We show that D-CDL is general and can be applied to multiple prompt-based CL methods, achieving consistent improvements over standard distillation baselines.
% \end{itemize}
In summary, our main contributions are as follows:
\begin{itemize}
    \item We introduce \textbf{Continual Distillation Learning (CDL)}, a distinct and practically important problem setting that studies distillation from larger models to smaller ones in rehearsal-free prompt-based continual learning.
    \item We analyze why existing distillation methods are suboptimal in prompt-based CDL, and identify two key challenges: objective entanglement in prompts and the lack of a persistent carrier for cross-task distillation.
    \item We propose \textbf{Decoupled Continual Distillation Learning (D-CDL)}, which introduces persistent KD-prompts to decouple task adaptation from distillation and enables global knowledge transfer across tasks.
    \item We validate D-CDL across four representative prompt-based continual learning methods, multiple datasets, and two teacher-student settings, showing consistent improvements over standard distillation baselines.
\end{itemize}

\section{Related Work}

\subsection{Continual Learning}

Continual learning (CL) aims to enable models to learn a sequence of tasks without catastrophically forgetting previously acquired knowledge~\citep{nguyen2019toward}. Previous CL methods can be broadly categorized into several families~\citep{wang2023Survey}. Regularization-based methods mitigate forgetting by constraining model updates during training. For example, function regularization methods~\citet{rebuffi2017icarl,li2016learning} preserve previous knowledge through distillation or output-level constraints, while weight regularization methods~\citet{kirkpatrick2017overcoming_EWC,zenke2017SI,liu2018rotate,ritter2018online,schwarz2018progress} penalize changes to important parameters based on their estimated contribution to past tasks. Architecture-based methods~\citet{yoon2017lifelong,jung2020continual,xu2018reinforced,fernando2017pathnet,rajasegaran2019random} address forgetting by dynamically allocating task-specific parameters or network components. Replay-based methods~\citet{rebuffi2017icarl,li2016learning,isele2018selective,lopezpaz2017gradient,castro2018end,chaudhry2019GEMA,chaudhry2019tinyMemory} alleviate forgetting by storing and replaying samples from previous tasks. More recently, increasing attention has been paid to rehearsal-free continual learning methods~\citet{hayes2019lifelong,gao2022rdfcil,Choi_2021_CVPR}, which aim to overcome catastrophic forgetting without using memory buffers and are therefore more desirable in privacy-sensitive and memory-constrained scenarios. Among different continual learning settings, rehearsal-free class-incremental learning is particularly challenging, since the model must continually learn new classes without access to previous data while being evaluated over all seen classes. This setting has therefore become an important benchmark for studying scalable and practical continual learning methods.

\subsection{Prompt-based Continual Learning}

With the rapid progress of large pre-trained vision models, prompt-based continual learning has emerged as a highly effective paradigm for rehearsal-free class-incremental learning. Instead of updating the entire backbone, these methods freeze a pre-trained Vision Transformer (ViT) and shift learning to a small number of trainable prompts, thereby preserving the strong pre-trained representation while reducing interference across tasks. 
Representative methods such as L2P~\citep{wang2022learning}, DualPrompt~\citep{wang2022dualprompt}, and CODA-Prompt~\citep{Smith2023CODAprompt} follow a prompt-pool paradigm, where the model selects the most relevant prompts or composes prompts from a prompt pool. CPrompt~\citep{gao2024cprompt} and OS-Prompt~\citep{kim2024osprompt} also build on the prompt-pool framework, while improving training-testing consistency and query efficiency, respectively. In contrast, DAP~\citep{jung2023dap} and APT~\citep{APT_2025_ICCV} remove prompt-pool and directly learn additive prompts. 
Despite these differences, prompt-based continual learning methods all rely heavily on the quality of the frozen pre-trained backbone, and larger ViT backbones often lead to better continual learning performance. This suggests that backbone scale is an important yet underexplored factor in rehearsal-free prompt-based continual learning, motivating a new setting where knowledge from a stronger backbone is distilled into a smaller one for improved efficiency and preserved continual learning performance.

% Despite their differences, these methods all rely heavily on the quality of the frozen pre-trained backbone. In particular, we observe that larger ViT backbones often lead to better continual learning performance in prompt-based CL. This suggests that backbone scale is an important but underexplored factor in rehearsal-free prompt-based continual learning. Such an observation motivates us to study a new problem setting, where knowledge from a stronger large backbone can be distilled into a smaller one to improve efficiency while preserving continual learning performance.
% Representative methods include L2P~\cite{wang2022learning}, which maintains a prompt pool and retrieves task-relevant prompts based on input features, and DualPrompt~\cite{wang2022dualprompt}, which further separates prompts into general prompts for task-invariant knowledge and expert prompts for task-specific adaptation. CODA-Prompt~\cite{Smith2023CODAprompt} departs from explicit key-prompt pair retrieval and instead constructs the final prompt through a weighted combination of prompt components. 
% More recent methods, such as Additive Prompt Tuning (APT)~\cite{APT_2025_ICCV}, abandon prompt-pool querying and directly learn a set of additive prompts that are injected into the self-attention process for class-incremental learning.

\subsection{Knowledge Distillation}

Knowledge distillation (KD) aims to transfer knowledge from a larger teacher model to a smaller student model so that the student can achieve strong performance with lower computational cost. Existing KD methods are typically divided into two main categories: logit distillation~\cite{hinton2015distilling,zhao2022decoupled}, which aligns the softened output distributions between teacher and student, and feature distillation~\cite{romero2014fitnets,chen2021ReviewKD,liang2023less}, which encourages the student to mimic intermediate representations of the teacher. KD has been widely studied in standard supervised learning, where both teacher and student are trained in a static task setting.

However, directly applying traditional KD methods to our continual distillation learning (CDL) setting is not straightforward. In prompt-based continual learning, the ViT backbone is frozen and adaptation is performed through task-dependent prompts, which creates a fundamentally different learning dynamics from conventional distillation scenarios. As a result, existing logit- and feature-based distillation methods yield only limited improvements in the CDL setting. This limitation motivates us to study how to perform knowledge distillation more effectively in prompt-based continual learning, leading to our proposed Decoupled Continual Distillation Learning (D-CDL).

\section{Continual Distillation Learning}

\subsection{Continual Learning Setting}
\label{sec:cl_setting}

In continual learning, a model is required to learn a sequence of tasks presented sequentially over time.
We denote a sequence of tasks as $\mathcal{D} = \{ \mathcal{D}_1, \ldots, \mathcal{D}_T \}$, where $T$ is the number of tasks.
The $t$th task $\mathcal{D}_t = \{ (\mathbf{x}_i^t, y_i^t)  \}_{i=1}^{n_t}$ consists of pairs of input sample $\mathbf{x}_i^t \in \mathcal{X}$ and its label $y_i^t \in \mathcal{Y}$, where $n_t$ is the number of samples for the $t$th task.
In this work, we consider rehearsal-free class-incremental learning, where each task consists of a fixed number of non-overlapping classes.
In training, a model learns these tasks one by one, and data from previous tasks are not available when training future tasks.
In testing, the model is evaluated on samples from all seen classes.

\subsection{Prompt-based Continual Learning Mechanism}
\label{sec:cl_mech}
Prompt-based continual learning methods typically build upon a pre-trained Vision Transformer (ViT) backbone and freeze the backbone parameters during training, while shifting task adaptation to learnable prompts.
A common view of these methods is that they maintain a prompt pool
$\mathcal{P}=\{\mathbf{p}_1,\mathbf{p}_2,\ldots,\mathbf{p}_M\}$,
where $\mathbf{p}_m$ denotes a prompt component and $M$ is the total number of prompt components.
Given an input image $\mathbf{x}$ from the current task, a query feature is first extracted by a query function $q(\mathbf{x})$, and then used to select or compose task-adaptive prompts from the prompt pool.
The resulting prompt $\mathbf{P}$ is injected into the frozen ViT backbone to assist classification.
Since each new task performs its own prompt construction again, the prompts used in prompt-based continual learning are essentially obtained through a \emph{task-specific selection} mechanism.
As a result, these prompts are not explicitly propagated to the next task; instead, only partial sharing is implicitly achieved through the prompt pool.
Meanwhile, because the backbone is frozen and each task mainly updates its own selected prompts, the interference across tasks is reduced, which helps alleviate catastrophic forgetting.

Representative prompt-based continual learning methods mainly differ in how the task-adaptive prompt $\mathbf{P}$ is constructed.
\textbf{L2P}~\citep{wang2022learning} associates each prompt component $\mathbf{p}_m$ with a learnable key $\mathbf{k}_m$.
For an input image $\mathbf{x}$, the cosine similarity between the query and each key is computed as $s_m=\gamma(q(\mathbf{x}),\mathbf{k}_m)$, where $\gamma(\cdot,\cdot)$ denotes cosine similarity.
Based on these similarity scores, L2P selects the top-$K$ prompt components from the prompt pool to form the final prompt for the current input:
$\mathbf{P}=\mathrm{TopK}(\mathcal{P}; \{s_m\}_{m=1}^{M})$.
The training objective is:
\begin{equation}
\min_{\mathcal{P}, \mathcal{K}, \phi}
\mathcal{L}\big(g_{\phi}(f_b(\mathbf{x})), y\big)
+
\lambda \sum_{\mathbf{k}_m \in \mathcal{K}_{\mathbf{x}}}
\gamma\!\left(q(\mathbf{x}), \mathbf{k}_m\right),
\label{eq:l2p_loss_short}
\end{equation}
where $\mathcal{K}=\{\mathbf{k}_m\}_{m=1}^{M}$ is the key set, $f_b$ is the ViT backbone, $g_{\phi}$ is the classifier, and $\mathcal{L}$ denotes the classification loss.

\textbf{DualPrompt}~\citep{wang2022dualprompt} extends the L2P framework by decomposing prompts into G-Prompts and E-Prompts, where G-Prompts capture task-shared knowledge and E-Prompts capture task-specific knowledge.
Moreover, these prompts can be inserted into multiple transformer blocks of the ViT backbone.

Instead of hard top-$K$ selection, \textbf{CODA-Prompt}~\citep{Smith2023CODAprompt} constructs the final prompt using a weighted combination of all prompt components, i.e.,
$\mathbf{P}=\sum_{m=1}^{M}\alpha_m \mathbf{p}_m$,
where the coefficient $\alpha_m$ is computed according to the similarity between the query feature and the corresponding key.
Specifically, CODA-Prompt introduces learnable feature-selection attention vectors
$\mathcal{A}=\{\mathbf{A}_1,\ldots,\mathbf{A}_M\}$ and computes the coefficient as
$\alpha_m=\gamma(q(\mathbf{x}) \odot \mathbf{A}_m,\mathbf{k}_m)$,
where $\odot$ denotes element-wise multiplication.
The final objective jointly optimizes the prompt components, keys, attention vectors, and classifier:
\begin{equation}
\min_{\mathcal{P}, \mathcal{K}, \mathcal{A}, \phi}
\mathcal{L}\big(g_{\phi}(f_b(\mathbf{x})), y\big)
+
\lambda \Big(
\mathcal{L}_{\mathrm{or}}(\mathcal{P})
+
\mathcal{L}_{\mathrm{or}}(\mathcal{K})
+
\mathcal{L}_{\mathrm{or}}(\mathcal{A})
\Big),
\label{eq:coda_loss_short}
\end{equation}
where $\mathcal{L}_{\mathrm{or}}$ denotes the orthogonality loss used to reduce interference between existing and newly learned knowledge.

More recent methods, such as Additive Prompt Tuning (\textbf{APT})~\citep{APT_2025_ICCV}, remove the prompt-pool querying mechanism and directly learn additive prompts for continual adaptation.
Despite this difference, the common goal of these prompt-based continual learning methods remains the same: shifting task adaptation from the frozen ViT backbone to a small set of learnable prompt parameters.
To ensure the diversity and generality of our evaluation, we build continual distillation experiments on top of four representative prompt-based continual learning methods: L2P, DualPrompt, CODA-Prompt, and APT.

\subsection{Prompting with Prefix-Tuning}

Given a prompt representation $\mathbf{P} \in \mathbb{R}^{L_p \times D}$, where $L_p$ denotes the prompt length and $D$ denotes the embedding dimension, we inject it into the ViT backbone for prompting.
Following prior prompt-based continual learning methods, we adopt the prefix-tuning~\citep{li2021prefix} scheme.
Specifically, $\mathbf{P}$ is divided into two parts, $\mathbf{P}_K$ and $\mathbf{P}_V \in \mathbb{R}^{L_p/2 \times D}$, which are appended to the key and value sequences, respectively, in the multi-head self-attention (MSA) layers:
\begin{equation}
f_{\mathrm{Pre\mbox{-}T}}(\mathbf{P}, \mathbf{h})
=
\mathrm{MSA}\!\left(
\mathbf{h}_Q,\,
[\mathbf{P}_K; \mathbf{h}_K],\,
[\mathbf{P}_V; \mathbf{h}_V]
\right),
\label{eq:PreT}
\end{equation}
where $\mathbf{h}_Q$, $\mathbf{h}_K$, and $\mathbf{h}_V$ denote the query, key, and value features in the attention layer, respectively, and $[\cdot;\cdot]$ denotes sequence concatenation. A key advantage of prefix-tuning is that it keeps the output token length unchanged, since the inserted prompts only modify the key and value branches of attention.
We employ prefix-tuning throughout this work to avoid introducing excessive additional tokens and computational overhead.

\subsection{Continual Distillation Learning Setup}

We consider the continual distillation learning (CDL) setup as illustrated in Fig.~\ref{fig:intro}(a), which follows the same continual learning setting described in Sec.~\ref{sec:cl_setting}.
For each task $t$, CDL maintains two prompt-based continual learners: a larger teacher model and a smaller student model.
The teacher model is first trained on the current task $\mathcal{D}_t$ and then frozen.
After that, the frozen teacher supervises the training of the student model on the same task.
Therefore, the teacher and student are optimized sequentially on each task under the same rehearsal-free class-incremental learning protocol.

In this work, to examine the effectiveness of existing distillation methods in the CDL setup, we build different types of knowledge distillation objectives, including logit distillation and feature distillation.
For logit distillation, the student is trained to match the teacher's output distribution at the classifier level.
For feature distillation, the student is trained to mimic the teacher's intermediate representations.
Since the ViT backbone consists of multiple transformer blocks, we use the internal tokens produced by each block as feature representations, allowing the student to learn from the teacher's internal token representations throughout the network.
The detailed formulations of the adopted logit distillation and feature distillation objectives are provided in the Appendix~\ref{appendix:Logit_KD} and ~\ref{appendix:Feature_KD}.

\section{Decoupled Continual Distillation Learning}

\subsection{Limitations of Existing Distillation in CDL}

Although previous knowledge distillation methods can be directly applied to the CDL setting, the performance gains are limited, and in some cases even negligible, as shown by the results in Fig.~\ref{fig:intro}.
This suggests that conventional distillation methods face fundamental limitations when applied to prompt-based continual distillation learning.

We argue that the primary issue arises from the prompt selection mechanism in prompt-based continual learning methods. In these methods, the ViT backbone must remain frozen during training (\emph{the necessity of freezing the backbone in prompt-based continual learning is also validated by the ablation study in our experiments Fig.~\ref{fig:unfreeze_task1} and Table~\ref{tab:unfreeze_ablation}}). As a result, for the student model, both the task-specific classification information of the current task and the knowledge distilled from the teacher can only be absorbed into the same prompts $(\mathbf{P}_K,\mathbf{P}_V)$, i.e., the green prompts in Fig.~\ref{fig:D-CDL}. These prompts then jointly influence the class token and the final prediction.

However, the prompts $(\mathbf{P}_K,\mathbf{P}_V)$ in prompt-based continual learning are obtained through a task-specific selection mechanism.
Due to the query-key based prompt construction process, the prompts used in each task are re-selected from the prompt pool according to the current task input, rather than being fully inherited from the prompts used in the previous task.
In other words, the initialization and optimization of $(\mathbf{P}_K,\mathbf{P}_V)$ at task $N$ do not completely come from the prompts of task $N\!-\!1$, but depend on a new round of prompt selection from the prompt pool.
Such a task-specific prompt selection mechanism is beneficial for continual learning, since it reduces interference between tasks and helps alleviate catastrophic forgetting.

Nevertheless, this property becomes problematic for knowledge distillation.
The knowledge learned from the teacher should not be interrupted together with the task-specific prompts across tasks, because the structural knowledge transferred from the teacher is expected to be useful globally across the whole task sequence, rather than only for a single task.
In other words, continual learning prefers task-specific prompts, while teacher distillation requires prompts that preserve task-shared and globally useful knowledge.
These two goals are inherently mismatched when they are entangled in the same $(\mathbf{P}_K,\mathbf{P}_V)$.
Therefore, in the CDL setting, the original prompts cannot effectively fulfill both continual adaptation and teacher-to-student distillation at the same time.

To address this limitation, we propose \textbf{Decoupled Continual Distillation Learning (D-CDL)}, as illustrated in Fig.~\ref{fig:D-CDL}.
Specifically, we keep the original prompts $(\mathbf{P}_K,\mathbf{P}_V)$ unchanged for continual learning, and introduce a new set of global KD-prompts $(\mathbf{P}_K^d,\mathbf{P}_V^d)$ that are specifically designed for teacher-to-student distillation across tasks.
We further introduce a dedicated KD-classifier to cooperate with the KD-prompts for distillation.
In this way, the prompting process is explicitly decoupled: the original prompts are still used to solve the continual learning problem, while the newly introduced KD-prompts and KD-classifier are dedicated to distillation.
This decoupled design separates the objectives of continual adaptation and knowledge distillation, enabling teacher knowledge to be preserved and propagated more effectively across tasks.

\subsection{Decoupled Prompting}

\begin{figure}
  \centering
  \includegraphics[width=1\linewidth]{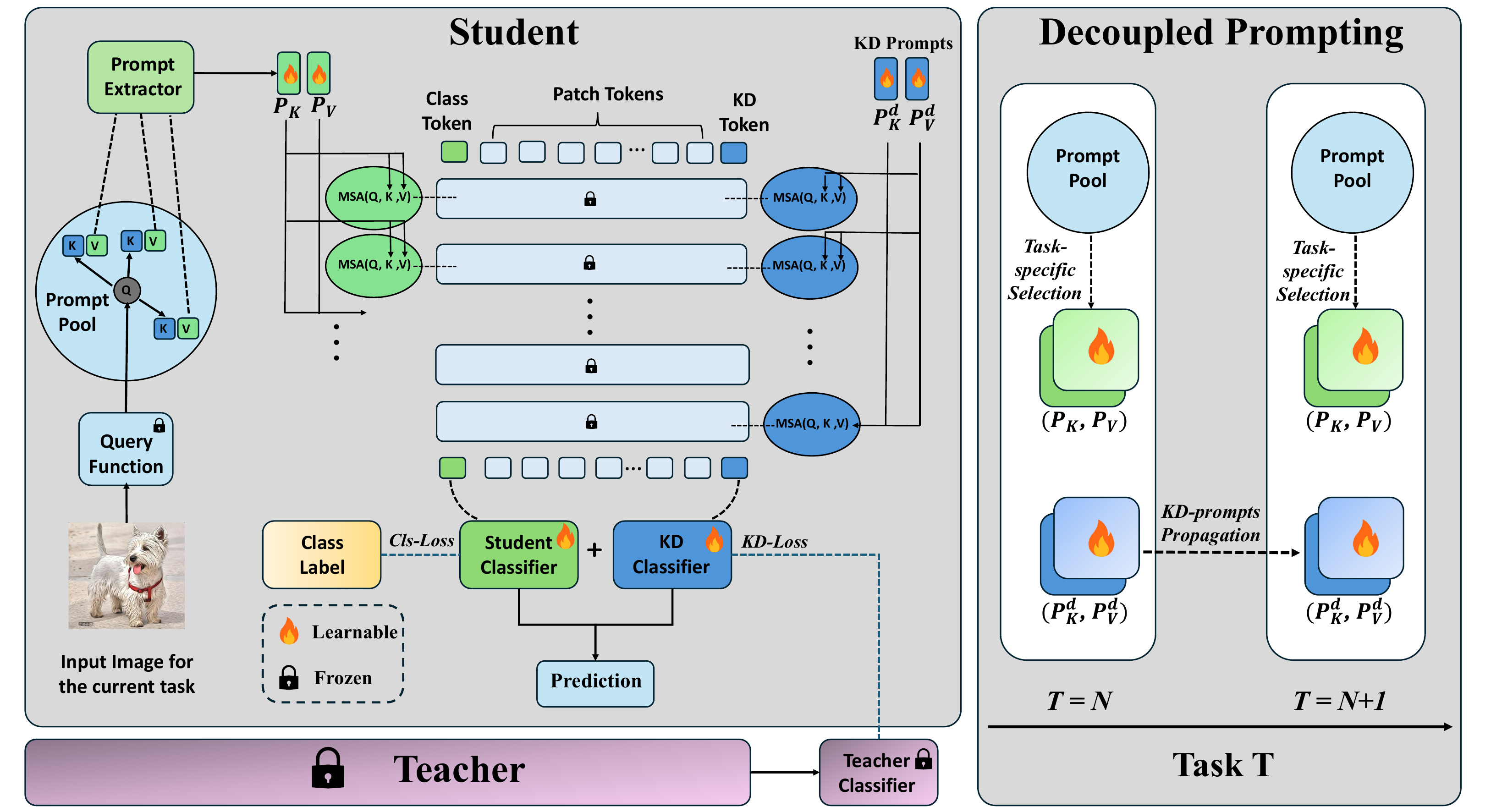}
  \caption{Overview of our Decoupled Continual Distillation Learning (D-CDL) framework. At each task, a teacher model is first trained on the current task and then frozen to provide distillation supervision for the student.
The student uses task-specific prompts $(P_K, P_V)$ for continual learning and separate KD-prompts $(P_K^d, P_V^d)$ for distilling teacher knowledge. \textbf{Right:} our decoupled prompting mechanism, where task-specific prompts are selected independently for each task, while KD-prompts are persistently propagated across tasks as a global carrier of distillation knowledge.
}

  \label{fig:D-CDL}
  %\vspace{-4mm}
\end{figure}

As illustrated in Fig.~\ref{fig:D-CDL}, an input image $\mathbf{x}$ is first processed by the query function to generate a query feature, which is then compared with the key set $\mathcal{K}$ in the prompt pool.
Following the prompt construction mechanism in Sec.~\ref{sec:cl_mech},  different prompt-based continual learners obtain continual-learning prompts through the prompt extractor, e.g., using top-$K$ selection in L2P or weighted combination in CODA-Prompt.
For methods without prompt-pool querying, such as APT, the continual-learning prompts are directly learned.
We denote the resulting prompts for continual learning by $\mathbf{P}$.
These prompts are then divided into $(\mathbf{P}_K,\mathbf{P}_V)$ and inserted into the MSA layers across multiple transformer blocks with prefix-tuning as in~\eqref{eq:PreT}, mainly serving the class token for the original classification objective.

Similar to the CDL setup, for each task the teacher model is first trained on the current task and then frozen to supervise the student.
Different from standard CDL, our D-CDL introduces a new set of KD-prompts together with a dedicated KD branch in the student model.
Similar to the original prompts $(\mathbf{P}_K,\mathbf{P}_V)$, the KD-prompts are also divided into $(\mathbf{P}_K^d,\mathbf{P}_V^d)$.
In addition, we append a new KD token to the end of the patch-token sequence at the input of the ViT backbone.
The KD-prompts $(\mathbf{P}_K^d,\mathbf{P}_V^d)$ are inserted into the MSA layers across multiple transformer blocks with prefix-tuning and are designed to primarily act on the newly introduced KD token.
Accordingly, the overall prefix-tuning operation in D-CDL is defined as
\begin{equation}
f_{\mathrm{Pre\mbox{-}T}}(\mathbf{P}, \mathbf{h}, \mathbf{P}^d)
=
\mathrm{MSA}\!\left(
\mathbf{h}_Q,\,
[\mathbf{P}_K;\mathbf{h}_K;\mathbf{P}_K^d],\,
[\mathbf{P}_V;\mathbf{h}_V;\mathbf{P}_V^d]
\right),
\label{eq:dcdl_pret}
\end{equation}
where $\mathbf{P}$ and $\mathbf{P}^d$ denote the continual-learning prompts and KD-prompts.

As illustrated in the right part of Fig.~\ref{fig:D-CDL}, D-CDL explicitly decouples the cross-task behaviors of the two prompt types.
The original prompts $(\mathbf{P}_K,\mathbf{P}_V)$ remain task-specific: they are selected or constructed independently for each task and are mainly responsible for continual learning.
In contrast, the KD-prompts $(\mathbf{P}_K^d,\mathbf{P}_V^d)$ are not re-selected in a task-specific manner.
Instead, they are persistently propagated across tasks, so that the teacher knowledge learned at the current task can be continuously preserved and transferred to future tasks.
In this way, D-CDL explicitly separates task-specific prompt adaptation from cross-task distillation knowledge propagation.

After being propagated through the ViT backbone, the final embedding of the KD token is fed into a separate classifier, termed the \emph{KD classifier}.
In the student model, this KD classifier is used to compute the distillation loss with the teacher classifier using the standard logit distillation objective.
That is, the logits produced by the teacher model serve as soft targets for the student, and the softened probability distributions are controlled by a temperature factor $\tau$.
The corresponding KD loss is defined as
\begin{equation}
\mathcal{L}_{\mathrm{KD}}
=
\tau^2 \sum_i p_i^{\mathcal{T}}
\log
\left(
\frac{p_i^{\mathcal{T}}}{p_i^{\mathcal{S}}}
\right),
\label{eq:kd_loss}
\end{equation}
where $p_i^{\mathcal{T}}$ and $p_i^{\mathcal{S}}$ denote the softened probability distributions of the teacher and student, respectively, and $\tau$ is the temperature parameter.

Meanwhile, the original student classifier only performs the standard classification objective with the ground-truth class label.
Therefore, the final training loss of the student model is
\begin{align}
\mathcal{L}_{\mathrm{S}}
&=
(1-\alpha)\,
\mathcal{L}
\big(
g_{\phi}^{\mathcal{S}}(f_{b}^{\mathcal{S}}(\mathbf{x};\mathbf{P},\mathbf{P}^d)),\, y
\big)
+
\alpha\,
\mathcal{L}_{\mathrm{KD}}
\big(
k_{\psi}^{\mathcal{S}}(f_{b}^{\mathcal{S}}(\mathbf{x};\mathbf{P},\mathbf{P}^d)),\,
g_{\phi}^{\mathcal{T}}(f_{b}^{\mathcal{T}}(\mathbf{x}))
\big)
+
\lambda \mathcal{L}_{\mathrm{pool}},
\label{eq:dcdl_loss}
\end{align}
where $g_{\phi}^{\mathcal{S}}$ and $k_{\psi}^{\mathcal{S}}$ denote the student classifier and KD classifier in the student model, respectively, and $g_{\phi}^{\mathcal{T}}$ denotes the teacher classifier.
$f_b^{\mathcal{S}}$ and $f_b^{\mathcal{T}}$ denote the frozen ViT backbones of the student and teacher models, respectively.
$\mathcal{L}_{\mathrm{pool}}$ denotes the loss associated with extracting continual-learning prompts from the prompt pool. For prompt-pool based methods, this term corresponds to the original prompt-pool related loss, as shown in the latter parts of~\eqref{eq:l2p_loss_short} and~\eqref{eq:coda_loss_short}, while for methods without prompt-pool querying, such as APT~\citep{APT_2025_ICCV}, this term is omitted.
$\alpha$ and $\lambda$ are balancing coefficients.

During testing, the outputs of the student classifier and the KD classifier are jointly used for prediction. Specifically, the final prediction is computed as a weighted combination of the two outputs, i.e., $(1-\alpha)$ for the student classifier and $\alpha$ for the KD classifier, where $\alpha$ is the same balancing coefficient used during training.
In this way, D-CDL explicitly decouples the two roles of prompting: $(\mathbf{P}_K,\mathbf{P}_V)$ are preserved for continual learning, while $(\mathbf{P}_K^d,\mathbf{P}_V^d)$ are dedicated to distilling and propagating teacher knowledge across tasks.

\section{Experiments}

\subsection{Experimental Setup}

% \textbf{Datasets:} We utilize the CIFAR-100~\citep{cifar100} and ImageNet-R~\citep{Hendrycks_2021_ICCV} datasets in a class-incremental continual learning setting for our experiments. Following previous works, we divide the ImageNet-R and CIFAR-100 datasets into 10 tasks, where each task contains 10 classes. 

\textbf{Datasets:} For the main experiments, we utilize the CIFAR-100~\citep{cifar100} and ImageNet-R~\citep{Hendrycks_2021_ICCV} datasets in a class-incremental continual learning setting for our experiments. Following previous works, we divide CIFAR-100 into 10 tasks with 10 classes per task, and ImageNet-R into 10 tasks with 20 classes per task.

\textbf{Implementation Details:}
We experiment with four prompt-based continual learning frameworks, including prompt-pool based methods, i.e., L2P~\citep{wang2022learning}, DualPrompt~\citep{wang2022dualprompt}, and CODA-Prompt~\citep{Smith2023CODAprompt}, as well as APT~\citep{APT_2025_ICCV}, which removes the prompt-pool selection mechanism and directly introduces additive prompts. The detailed experimental settings of the four prompt-based continual learning methods are provided in the Appendix~\ref{appendix:CL_experimental_details}.
For continual distillation, the teacher and student adopt the same prompt-based continual learning framework, while the teacher uses a larger ViT backbone.
To improve the generality of our conclusions, we consider two teacher-student distillation settings: ViT-Large to ViT-Base and ViT-Base to ViT-Small.
For prompt-pool based methods, the coefficient $\lambda$ associated with $\mathcal{L}_{\mathrm{pool}}$ is set to $1$.
For all distillation methods, the temperature parameter is set to $\tau = 2$.
In our D-CDL method, KD-prompts are inserted into every transformer block, and the lengths of $\mathbf{P}_K^d$ and $\mathbf{P}_V^d$ in each block are both set to $3$.
The balancing parameter $\alpha$ in~\eqref{eq:dcdl_loss} is set to $0.5$.

\textbf{Training Details:}
For all experiments, we use the Adam optimizer~\citep{kingma2014adam}.
The Split ImageNet-R dataset is trained for $35$ epochs, while the Split CIFAR-100 dataset is trained for $20$ epochs.
The batch size is set to $64$ for all experiments, and the learning rate is fixed at $0.001$.
All experiments are conducted on two NVIDIA RTX 6000 Ada GPUs with 48GB memory.
The total student training time over all tasks is about 4 hours on Split CIFAR-100 and 3 hours on Split ImageNet-R for ViT-Base students, and about 2 hours on Split CIFAR-100 and 1.5 hours on Split ImageNet-R for ViT-Small students.

\textbf{Evaluation Metrics:} Our experiments use two metrics to evaluate the models: accuracy and forgetting rate~\citep{lopezpaz2017gradient}. Accuracy refers to the average accuracy of all tasks after completing all 10 tasks. The average forgetting rate reflects the influence of learning a new task on previously completed tasks. A higher value signifies a more negative impact of the continual learning model. We provide the details in Appendix~\ref{appendix:Metrics}.

\subsection{Distillation Results}

\begin{table}[t]
\centering
\vspace{-4pt}
\caption{
Continual distillation results on the ImageNet-R dataset with ViT-Small students (ViT-Base teachers) under different prompt-based continual learning frameworks.
}
\vspace{-4pt}
\label{tab:imagenetr_small_all_methods}
\renewcommand{\arraystretch}{1.0}
\resizebox{\textwidth}{!}{
\begin{tabular}{l|cc|cc|cc|cc}
\toprule
\multirow{2}{*}{Methods}
& \multicolumn{2}{c|}{L2P~\citep{wang2022learning}}
& \multicolumn{2}{c|}{DualPrompt~\citep{wang2022dualprompt}}
& \multicolumn{2}{c|}{CODA-Prompt~\citep{Smith2023CODAprompt}}
& \multicolumn{2}{c}{APT~\citep{APT_2025_ICCV}} \\
& Avg. Acc ($\uparrow$) & Forgetting ($\downarrow$)
& Avg. Acc ($\uparrow$) & Forgetting ($\downarrow$)
& Avg. Acc ($\uparrow$) & Forgetting ($\downarrow$)
& Avg. Acc ($\uparrow$) & Forgetting ($\downarrow$) \\
\midrule
Baseline (w/o Distillation)

& 63.82 $\pm$ 0.25 & 6.52 $\pm$ 0.31
& 65.51 $\pm$ 0.11 & 5.93 $\pm$ 0.03
& 67.44 $\pm$ 0.46 & 8.52 $\pm$ 0.05
& 71.71 $\pm$ 0.17 & 4.25 $\pm$ 0.04 \\
\midrule
CDL-KD~\citep{hinton2015distilling}
& 63.97 $\pm$ 0.62 & 6.51 $\pm$ 0.06
& 65.68 $\pm$ 0.06 & 7.26 $\pm$ 0.29
& 69.91 $\pm$ 0.62 & 7.64 $\pm$ 0.71
& 72.65 $\pm$ 0.07 & 3.92 $\pm$ 0.30 \\
CDL-DKD~\citep{zhao2022decoupled}
& 62.91 $\pm$ 0.27 & 6.55 $\pm$ 0.17
& 65.44 $\pm$ 0.05 & 7.27 $\pm$ 0.23
& 68.92 $\pm$ 0.07 & 8.39 $\pm$ 0.36
& 71.59 $\pm$ 0.51 & 3.90 $\pm$ 0.22 \\
CDL-FitNets~\citep{romero2014fitnets}
& 64.29 $\pm$ 0.09 & 6.37 $\pm$ 0.17
& 66.20 $\pm$ 0.14 & 5.93 $\pm$ 0.17
& 69.87 $\pm$ 0.04 & 7.38 $\pm$ 0.36
& 72.75 $\pm$ 0.07 & 4.03 $\pm$ 0.10 \\
CDL-ReviewKD~\citep{chen2021ReviewKD}
& 63.64 $\pm$ 0.34 & 6.36 $\pm$ 0.58
& 65.69 $\pm$ 0.91 & 6.56 $\pm$ 0.46
& 70.19 $\pm$ 0.16 & 7.68 $\pm$ 0.01
& 72.70 $\pm$ 0.14 & 3.93 $\pm$ 0.04 \\
CDL-DeiT~\citep{DeiT2020}
& 64.99 $\pm$ 0.49 & 3.83 $\pm$ 0.85
& 65.82 $\pm$ 0.48 & 4.00 $\pm$ 0.19
& 70.74 $\pm$ 0.20 & 6.66 $\pm$ 0.28
& 72.34 $\pm$ 0.27 & \textbf{3.71 $\pm$ 0.20} \\
\midrule
\textbf{D-CDL (ours)}
& \textbf{68.18 $\pm$ 0.03} & \textbf{2.08 $\pm$ 0.28}
& \textbf{68.77 $\pm$ 0.16} & \textbf{3.13 $\pm$ 0.25}
& \textbf{71.92 $\pm$ 0.50} & \textbf{5.61 $\pm$ 0.34}
& \textbf{73.29 $\pm$ 0.12} & 3.73 $\pm$ 0.06 \\
\midrule
\bottomrule
\end{tabular}
}
\end{table}
% \vspace{-20pt}

Table~\ref{tab:imagenetr_small_all_methods} summarizes the continual distillation results on the ImageNet-R dataset with ViT-Small students and ViT-Base teachers under different prompt-based continual learning frameworks.
Among the compared distillation methods, KD~\citep{hinton2015distilling} and DKD~\citep{zhao2022decoupled} are logit distillation methods, while FitNets~\citep{romero2014fitnets} and ReviewKD~\citep{chen2021ReviewKD} are feature distillation methods.
DeiT~\citep{DeiT2020} is a distillation method applicable to transformer architectures.
All these methods are adapted to the continual distillation learning setting in our experiments.
From the results, we can see that our proposed D-CDL consistently achieves better performance than existing distillation methods across all four prompt-based continual learning frameworks.
In particular, D-CDL obtains the highest average accuracy in all cases, and its forgetting is also the lowest or among the lowest in most settings.
These results indicate that D-CDL more effectively addresses the new problem of performing knowledge distillation in continual learning. Another important observation is that the effectiveness of D-CDL is not limited to prompt-pool based continual learning methods such as L2P~\citep{wang2022learning}, DualPrompt~\citep{wang2022dualprompt}, and CODA-Prompt~\citep{Smith2023CODAprompt}.
It also generalizes well to the more recent APT~\citep{APT_2025_ICCV} framework, which abandons the prompt-pool based task-specific selection mechanism and directly learns additive prompts.
This demonstrates that the benefit of D-CDL mainly comes from the proposed decoupled design, which explicitly separates continual adaptation from teacher-to-student distillation.

\begin{table}[t]
\centering
\vspace{-2pt}
\caption{
Comparison between Baseline (w/o distillation) and \textbf{D-CDL (ours)} under four prompt-based continual learning frameworks on the Split CIFAR-100 and Split ImageNet-R datasets. Results are reported under two teacher-student settings: ViT-Large $\rightarrow$ ViT-Base and ViT-Base $\rightarrow$ ViT-Small.
}
\vspace{-4pt}
\label{tab:dcdl_generality}
\resizebox{\textwidth}{!}{%
\begin{tabular}{ll|cc|cc|cc|cc}
\toprule
\multirow[b]{3}{*}{CL Framework} & \multirow[b]{3}{*}{Methods}
& \multicolumn{4}{c|}{\textbf{Split CIFAR-100}}
& \multicolumn{4}{c}{\textbf{Split ImageNet-R}} \\
\cmidrule(lr){3-6} \cmidrule(lr){7-10}
& 
& \multicolumn{2}{c|}{\textbf{ViT-Large $\rightarrow$ ViT-Base}}
& \multicolumn{2}{c|}{\textbf{ViT-Base $\rightarrow$ ViT-Small}}
& \multicolumn{2}{c|}{\textbf{ViT-Large $\rightarrow$ ViT-Base}}
& \multicolumn{2}{c}{\textbf{ViT-Base $\rightarrow$ ViT-Small}} \\
& 
& Avg. Acc ($\uparrow$) & Forgetting ($\downarrow$)
& Avg. Acc ($\uparrow$) & Forgetting ($\downarrow$)
& Avg. Acc ($\uparrow$) & Forgetting ($\downarrow$)
& Avg. Acc ($\uparrow$) & Forgetting ($\downarrow$) \\
\midrule
\multirow{2}{*}{L2P~\citep{wang2022learning}}
& Baseline
& 83.02 $\pm$ 0.47 & 6.06 $\pm$ 0.47
& 77.71 $\pm$ 0.49 & 7.12 $\pm$ 0.33
& 73.94 $\pm$ 0.22 & 4.41 $\pm$ 0.18
& 63.82 $\pm$ 0.25 & 6.52 $\pm$ 0.31 \\
& \textbf{D-CDL (ours)}
& \textbf{86.56 $\pm$ 0.22} & \textbf{4.97 $\pm$ 0.07}
& \textbf{81.79 $\pm$ 0.66} & \textbf{4.31 $\pm$ 0.27}
& \textbf{76.91 $\pm$ 0.40} & \textbf{3.15 $\pm$ 0.39}
& \textbf{68.18 $\pm$ 0.03} & \textbf{2.08 $\pm$ 0.28} \\
\midrule
\multirow{2}{*}{DualPrompt~\citep{wang2022dualprompt}}
& Baseline
& 84.66 $\pm$ 0.87 & 5.91 $\pm$ 0.34
& 79.85 $\pm$ 0.57 & 6.12 $\pm$ 0.32
& 73.18 $\pm$ 0.33 & \textbf{3.45 $\pm$ 0.32}
& 65.51 $\pm$ 0.11 & 5.93 $\pm$ 0.03 \\
& \textbf{D-CDL (ours)}
& \textbf{86.92 $\pm$ 0.24} & \textbf{4.77 $\pm$ 0.58}
& \textbf{81.78 $\pm$ 0.17} & \textbf{3.63 $\pm$ 0.03}
& \textbf{76.06 $\pm$ 0.12} & 3.77 $\pm$ 0.38
& \textbf{68.77 $\pm$ 0.16} & \textbf{3.13 $\pm$ 0.25} \\
\midrule
\multirow{2}{*}{CODA-Prompt~\citep{Smith2023CODAprompt}}
& Baseline
& 86.16 $\pm$ 0.17 & 5.63 $\pm$ 0.25
& 82.18 $\pm$ 0.20 & 6.48 $\pm$ 0.48
& 76.42 $\pm$ 0.17 & 4.31 $\pm$ 0.18
& 67.44 $\pm$ 0.46 & 8.52 $\pm$ 0.05 \\
& \textbf{D-CDL (ours)}
& \textbf{87.13 $\pm$ 0.09} & \textbf{5.30 $\pm$ 0.06}
& \textbf{84.31 $\pm$ 0.01} & \textbf{5.63 $\pm$ 0.02}
& \textbf{78.62 $\pm$ 0.57} & \textbf{3.46 $\pm$ 0.53}
& \textbf{71.92 $\pm$ 0.50} & \textbf{5.61 $\pm$ 0.34} \\
\midrule
\multirow{2}{*}{APT~\citep{APT_2025_ICCV}}
& Baseline
& 88.26 $\pm$ 0.09 & 3.31 $\pm$ 0.33
& 83.30 $\pm$ 0.15 & 4.35 $\pm$ 0.09
& 78.67 $\pm$ 0.28 & 4.06 $\pm$ 0.29
& 71.71 $\pm$ 0.17 & 4.25 $\pm$ 0.04 \\
& \textbf{D-CDL (ours)}
& \textbf{89.41 $\pm$ 0.04} & \textbf{2.74 $\pm$ 0.09}
& \textbf{84.28 $\pm$ 0.11} & \textbf{3.89 $\pm$ 0.09}
& \textbf{79.83 $\pm$ 0.19} & \textbf{3.87 $\pm$ 0.19}
& \textbf{73.29 $\pm$ 0.12} & \textbf{3.73 $\pm$ 0.06} \\
\midrule
\bottomrule
\end{tabular}%
}
\vspace{-8pt}
\end{table}

Table~\ref{tab:dcdl_generality} summarizes the comparison between the Baseline (without distillation) and our proposed D-CDL under four prompt-based continual learning frameworks, two datasets, and two teacher-student settings. Overall, D-CDL consistently improves the average accuracy over the baseline in all cases, demonstrating both the diversity and generality of our method across different continual learning frameworks and teacher-student settings. This shows that the proposed decoupled prompting design is not tied to a specific continual learning framework, dataset, or model scale. These results indicate that D-CDL provides a robust and general solution to continual distillation learning, rather than a gain limited to a particular benchmark or prompt construction strategy. 
More specifically, the APT model distilled by D-CDL achieves the strongest performance among the evaluated methods.
The complete comparison with other distillation methods is provided in the Appendix~\ref{appendix:Addtional_results}.

To further evaluate the generalization ability of D-CDL beyond the two datasets considered in the main experiments, we additionally conduct experiments on CUB-200~\citep{wah2011caltech}, a fine-grained visual recognition benchmark comprising 200 bird species. Detailed benchmark settings and experimental results are provided in Appendix~\ref{appendix:cub200}.

\subsection{Training and Parameter Overhead}

To further analyze the computational and parameter overhead introduced by D-CDL, we report the student training time and the additional parameter size of the proposed KD-prompt branch. The timing experiment is conducted on Split ImageNet-R under the CODA-Prompt framework with the ViT-Base $\rightarrow$ ViT-Small distillation setting. We compare D-CDL with the baseline without distillation and the standard logit distillation baseline CDL-KD~\citep{hinton2015distilling}. All reported times are measured in seconds and correspond to the total student training time over all 10 tasks.

\begin{table}[t]
\centering
\caption{Student training time comparison among the baseline without distillation, CDL-KD~\citep{hinton2015distilling}, and D-CDL on Split ImageNet-R under CODA-Prompt with the ViT-Base $\rightarrow$ ViT-Small distillation setting.}
\label{tab:training_time_overhead}
{
\resizebox{0.65\linewidth}{!}{
\begin{tabular}{l c c}
\toprule
Method & Student Training Time & Relative Cost \\
\midrule
Baseline (w/o Distillation) & $\sim$4300s & 1.00$\times$ \\
CDL-KD~\citep{hinton2015distilling} & $\sim$6100s & $\sim$1.42$\times$ \\
D-CDL (ours) & $\sim$6200s & $\sim$1.44$\times$ \\
\bottomrule
\end{tabular}
}
}
\end{table}

As shown in Table~\ref{tab:training_time_overhead}, D-CDL introduces only a small training-time overhead compared with CDL-KD~\citep{hinton2015distilling}. This indicates that the proposed decoupled KD-prompt branch adds only minor extra training cost compared with standard continual distillation.

\begin{table}[t]
\centering
\caption{Additional parameter overhead introduced by D-CDL to the student model in the ViT-Base $\rightarrow$ ViT-Small distillation setting.}
\label{tab:param_overhead}
{
\resizebox{0.65\linewidth}{!}{
\begin{tabular}{l c c}
\toprule
Component & Number of Parameters & Ratio to Backbone \\
\midrule
Backbone & $\sim$22M & 100\% \\
Additional D-CDL parameters & $\sim$0.1M & $\sim$0.5\% \\
\bottomrule
\end{tabular}
}
}
\end{table}

As shown in Table~\ref{tab:param_overhead}, the additional parameters introduced by D-CDL, including the KD-token, KD-prompts, and KD-classifier, are lightweight compared with the ViT-Small backbone. Therefore, the gains of D-CDL are achieved with very small parameter and training-time overhead.

\subsection{Ablation Studies}

\textbf{Necessity of Freezing the ViT Backbone:}
Our analysis of the limitations of conventional distillation in the CDL setting is built upon a frozen ViT backbone, which is also a common prerequisite of prompt-based continual learning methods. Since prompt-based CL shifts continual learning to task-specific prompts by freezing the pre-trained ViT backbone, a natural question is whether distillation can be improved by simply unfreezing part of the backbone.

\begin{figure}[t]
\centering

%==================== Row 1: two figures ====================%
\noindent
\begin{minipage}[t]{0.45\linewidth}
\vspace{0pt}
\centering
\includegraphics[width=\linewidth]{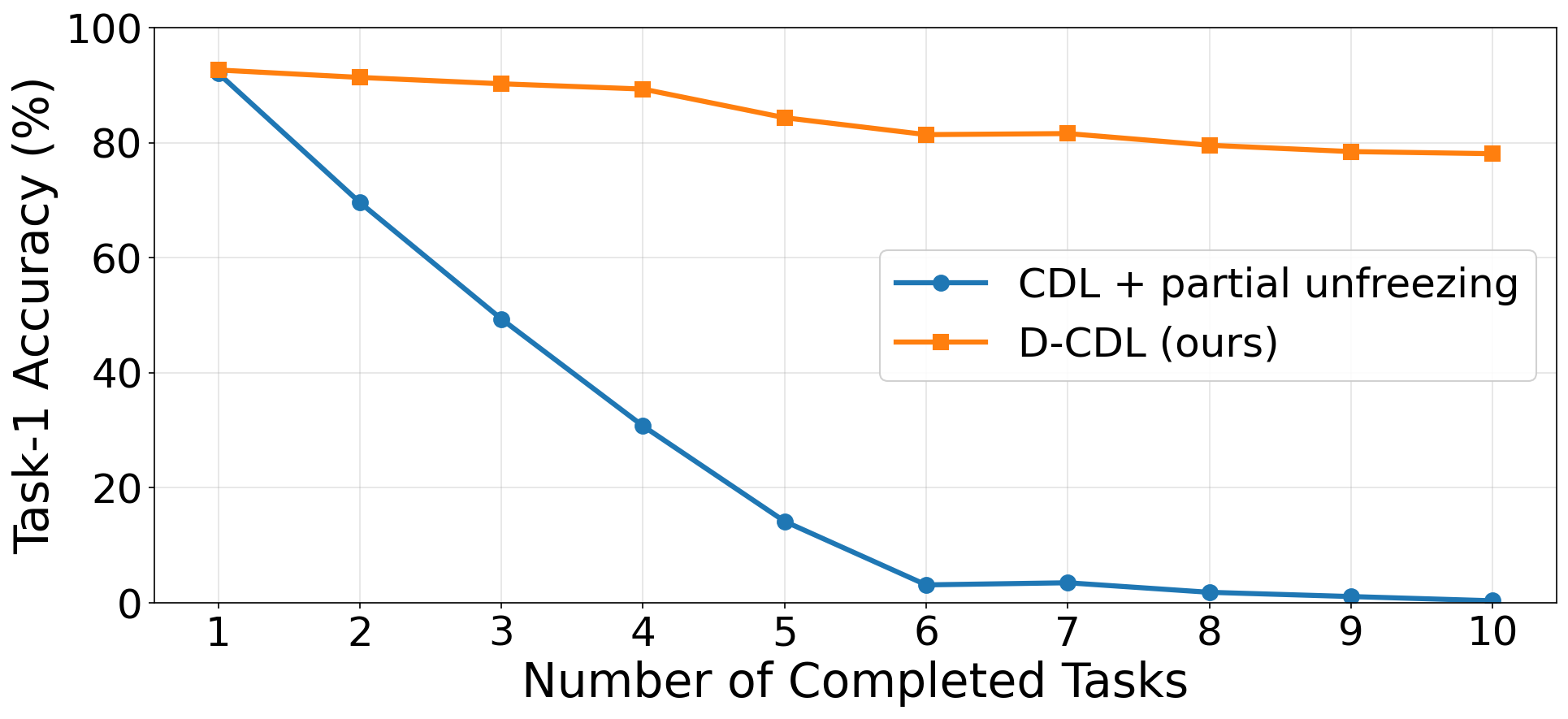}
\vspace{-14pt}

\caption{Task-1 accuracy after completing different numbers of tasks. CODA-Prompt with the ViT-Base $\rightarrow$ ViT-Small distillation on Split ImageNet-R}
\label{fig:unfreeze_task1}
\end{minipage}\hspace{0.09\linewidth}
\begin{minipage}[t]{0.45\linewidth}
\vspace{0pt}
\centering
\includegraphics[width=\linewidth]{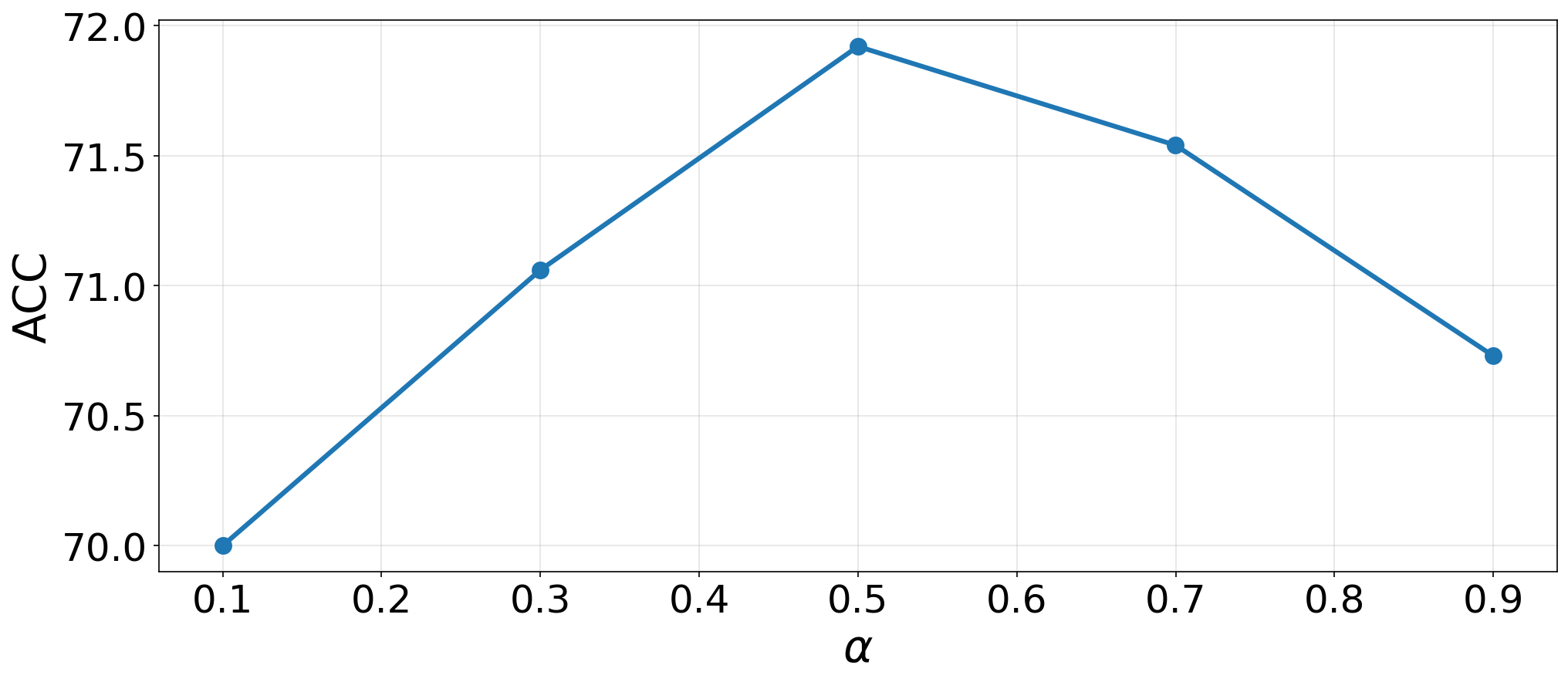}
\vspace{-14pt}
\caption{Effect of the balancing parameter $\alpha$ on average accuracy. CODA-Prompt with the ViT-Base $\rightarrow$ ViT-Small distillation on Split ImageNet-R}
\label{fig:alpha_ablation}
\end{minipage}

\vspace{1em}

%==================== Row 2: two tables ====================%
\begin{minipage}[t]{0.45\linewidth}
\vspace{0pt}
\centering
\captionof{table}{Final performance after all 10 tasks. CODA-Prompt with the ViT-Base $\rightarrow$ ViT-Small distillation on Split ImageNet-R}
\vspace{-4pt}
\label{tab:unfreeze_ablation}
% \vspace{0.1em}

\resizebox{\linewidth}{!}{
\begin{tabular}{lcc}
\toprule
Method & Avg. Acc. ($\uparrow$) & Forgetting ($\downarrow$) \\
\midrule
Baseline & 67.44 & 8.52 \\
CDL (unfreezing) & 30.06 & 58.24 \\
\textbf{D-CDL (ours)} & \textbf{71.92} & \textbf{5.61} \\
\bottomrule
\end{tabular}
}
\end{minipage}\hspace{0.09\linewidth}
\begin{minipage}[t]{0.45\linewidth}
\vspace{0pt}
\centering
\captionof{table}{Effect of the components in D-CDL. CODA-Prompt with the ViT-Base $\rightarrow$ ViT-Small distillation on Split ImageNet-R}
\vspace{-4pt}
\label{tab:kd_classifier_global}
% \vspace{0.1em}

\resizebox{0.9\linewidth}{!}{
\begin{tabular}{ccc}
\toprule
KD-Classifier & KD-prompts Propagation & Avg. Acc. ($\uparrow$) \\
\midrule
\multicolumn{2}{c}{\hspace{0em}\textit{Baseline}} & 67.44 \\
$\varnothing$ & $\checkmark$ & 70.21 \\
$\checkmark$ & $\varnothing$ & 71.02 \\
$\checkmark$ & $\checkmark$ & \textbf{71.92} \\
\bottomrule
\end{tabular}
}
\end{minipage}
\vspace{-8pt}
\end{figure}

To answer this question, we unfreeze the last transformer block of the student backbone, while keeping the other settings the same as in D-CDL. As shown in Fig.~\ref{fig:unfreeze_task1}, partially unfreezing the backbone causes severe forgetting: the first task is almost completely forgotten after all 10 tasks. Table~\ref{tab:unfreeze_ablation} further shows that this strategy substantially degrades the final average accuracy and increases forgetting. These results indicate that freezing the ViT backbone is necessary in prompt-based CL. At the same time, this also highlights the core difficulty of CDL: continual learning relies on task-specific prompts, whereas distillation requires knowledge to be preserved globally across tasks. This observation further motivates our decoupled prompting design.

\textbf{Effect of the Balancing Parameter $\boldsymbol{\alpha}$:}
Fig.~\ref{fig:alpha_ablation} shows the effect of the balancing $\alpha$, which is used both in the training objective in~\eqref{eq:dcdl_loss} and in the final prediction during test-time inference. The average accuracy first increases as $\alpha$ becomes larger, reaches the best performance at $\alpha=0.5$, and then gradually decreases. This result indicates that the best performance is achieved when the two parts in the decoupled prompting process are balanced equally.

\textbf{Effect of the components in D-CDL:}
Table~\ref{tab:kd_classifier_global} presents the contribution of different components in D-CDL. \emph{KD-classifier} denotes whether a dedicated KD-classifier is introduced. Without it, the KD-prompts are directly applied to the original class token, without forming a separate distillation branch. \emph{KD-prompts Propagation} denotes whether the KD-prompts of the current task are inherited from the previous task. When disabled, the KD-prompts are re-initialized for every new task. The results show that the KD-classifier plays an important role in the decoupled design, while KD-prompts propagation is also critical for effective distillation. This suggests that teacher knowledge should be preserved globally across tasks, and the KD-prompts should be continuously inherited throughout the continual learning process.

\textbf{Mechanistic Analysis:}
To further understand how D-CDL realizes decoupled distillation, we analyze the teacher alignment of the two student branches and the cross-task stability of the prompt parameters. The results show that the KD branch is more closely aligned with the teacher output than the original CLS branch, while the KD-prompts remain substantially more stable across tasks than the task-adaptive CL prompts. These observations provide further evidence that the KD branch serves as a dedicated pathway for teacher knowledge transfer and that KD-prompts act as a persistent carrier of distillation knowledge across tasks. Detailed analyses are provided in Appendix~\ref{appendix:mechanistic_analysis}.

% In addition, we provide a further analysis of the KD-prompt length and the number of inserted layers in Appendix~\ref{appendix:Layers_Length}.

\textbf{Additional Ablation Studies:}
We provide further analyses of the KD-prompt design in the appendix. 
Appendix~\ref{appendix:Layers_Length} investigates the effects of the 
KD-prompt length and the number of transformer layers into which the 
KD-prompts are inserted. Appendix~\ref{appendix:more_ablation} presents 
parameter-matched and two-head controlled baselines, together with 
ablations on KD-prompt attachment and cross-task propagation. These 
results further demonstrate that the improvements of D-CDL cannot be 
explained solely by additional prompt parameters or two-head ensembling, 
and support the importance of both the decoupled KD-prompt pathway and 
cross-task KD-prompt propagation.

\vspace{-8pt}

\section{Conclusion}

% We introduced \textbf{Continual Distillation Learning (CDL)}, a new problem setting for knowledge distillation in rehearsal-free prompt-based continual learning. We showed that conventional distillation methods are limited in this setting due to objective entanglement in task-specific prompts and the lack of cross-task continuity for distillation knowledge. To address this limitation, we proposed \textbf{Decoupled Continual Distillation Learning (D-CDL)}, which introduces persistent KD-prompts and a dedicated KD branch to explicitly decouple continual adaptation from distillation. Experiments on the Split CIFAR-100 dataset and the Split ImageNet-R dataset across four representative prompt-based continual learning methods showed that D-CDL consistently improves over standard distillation baselines. We hope this work will inspire future research on continual distillation in prompt-based continual learning.

We introduced Continual Distillation Learning (CDL), a new problem
setting for knowledge distillation in rehearsal-free prompt-based
continual learning. We showed that conventional distillation methods
are limited in this setting due to objective entanglement in task-specific
prompts and the lack of cross-task continuity for distillation knowledge.
To address this limitation, we proposed Decoupled Continual Distillation
Learning (D-CDL), which introduces persistent KD-prompts and a dedicated
KD branch to explicitly decouple continual adaptation from distillation.
Experiments across multiple datasets, four representative prompt-based
continual learning methods, and two teacher-student settings showed that
D-CDL consistently improves over standard distillation baselines. We hope
this work will inspire future research on continual distillation in
prompt-based continual learning.

\section*{ACKNOWLEDGMENT}
This work was supported in part by the National Science Foundation (NSF) under Grant Nos. 2346528 and 2520553, and the NVIDIA Academic Grant Program Award, and a gift funding from XPeng.

\clearpage

\bibliography{collas2026_conference}
\bibliographystyle{collas2026_conference}

\appendix

\section{Appendix}

\subsection{Relationship Between ViT Backbone Size and Performance}
\label{appendix:ViT_size}

\vspace{-2mm}
\begin{table*}[h]
    \centering
        \caption{Comparison of accuracy on the continual learning methods using different sizes of ViT backbones. 
}
    \scalebox{0.8}{
        \begin{tabular}{l c }
        \toprule
        \textbf{Model} & \textbf{Accuracy}  \\
      
        \midrule
        L2P~\citep{wang2022learning}-Tiny       & 60.68  \\
        L2P~\citep{wang2022learning}-Small      & 77.71   \\
        L2P~\citep{wang2022learning}-Base       & 83.02  \\
        L2P~\citep{wang2022learning}-Large      & \textbf{86.36}   \\
        \midrule
        CODA-Prompt~\citep{Smith2023CODAprompt}-Tiny       & 65.05  \\
        CODA-Prompt~\citep{Smith2023CODAprompt}-Small      & 82.18   \\
        CODA-Prompt~\citep{Smith2023CODAprompt}-Base       & 86.16   \\
        CODA-Prompt~\citep{Smith2023CODAprompt}-Large      & \textbf{88.97}  \\    
        \bottomrule
    \end{tabular}
    }
    \vspace{-1mm}
    \label{tab:model_comparison}
\end{table*}
\vspace{-7pt}

On the Split CIFAR-100 dataset with 10 continual learning tasks, we observe that the performance of prompt-based continual learning methods is strongly correlated with the quality of the pre-trained ViT backbone, which is largely determined by its model size. Specifically, larger ViT backbones consistently yield better continual learning performance.

\subsection{Detailed Experimental Settings of the Prompt-based Continual Learning Methods}
\label{appendix:CL_experimental_details}
In L2P, the prompt pool consists of a total of 30 prompt components, and a CL prompt with a length of 20 is inserted only in the first layer of the ViT backbone. In DualPrompt, the prompt pool contains 10 prompt components, each CL prompt has a length of 20, with G-prompts and E-prompts placed in the first two blocks and the third to fifth blocks, respectively. In CODA-Prompt, the prompt pool consists of 100 prompt components. Each CL prompt has a length of 8, and CL prompts are inserted from the first to the fifth layer of the ViT. 

In APT, there is no prompt pool. Instead, a single set of prompts is shared across all tasks. Specifically, each transformer layer is equipped with a pair of learnable additive prompts, one for the key and one for the value, which are directly added to the key and value vectors of the CLS token during self-attention. In addition, APT adopts a Progressive Prompt Fusion (PPF) strategy during inference, where the prompts from the previous task and the newly learned prompts are fused as $P_{t+1}^{\mathrm{PPF}}=\beta P_t + (1-\beta) P_{t+1}$, with $\beta \in [0,1]$ controlling the balance between old and new prompts. In our experiments, the fusion coefficient is set to $\beta = 0.7$.

\subsection{Evaluation Metrics}
\label{appendix:Metrics}

Our experiments use two metrics to evaluate the models: accuracy and forgetting rate~\citep{lopezpaz2017gradient}. Accuracy refers to the average accuracy of all tasks after completing all 10 tasks, defined by ~\eqref{eq:ACC}, where $R_{i,j}$ is the test classification accuracy of the model on task $t_{j}$ after observing the last sample from task $t_{i}$. $T$ is the total number of tasks. The forgetting rate (~\eqref{eq:Forgetting}) reflects the influence of learning a new task on previously completed tasks. A higher value signifies a more negative impact of the continual learning model.
\begin{equation}
\text{ACC} = \frac{1}{T} \sum_{i=1}^{T} R_{T,i},
\label{eq:ACC}
\end{equation}
\begin{equation}
\text{Forgetting} = \frac{1}{T-1} \sum_{i=1}^{T-1} (R_{i,i} - R_{T,i}).
\label{eq:Forgetting}
\end{equation}

\subsection{Additional Continual Distillation Results}
\label{appendix:Addtional_results}
For completeness, we provide the full continual distillation results under the remaining three teacher-student settings in Tables~\ref{tab:imagenetr_base_all_methods}, \ref{tab:cifar100_small_all_methods}, and \ref{tab:cifar100_base_all_methods}. The results further show that D-CDL consistently achieves the best or highly competitive performance across different prompt-based continual learning frameworks and distillation settings.

% Table #######################

\begin{table}[!h]
\centering
\caption{
Continual distillation results on the ImageNet-R dataset with ViT-Base students (ViT-Large teachers) under different prompt-based continual learning frameworks.
}
\label{tab:imagenetr_base_all_methods}
\renewcommand{\arraystretch}{1.0}
\resizebox{\textwidth}{!}{
\begin{tabular}{l|cc|cc|cc|cc}
\toprule
\multirow{2}{*}{Methods}
& \multicolumn{2}{c|}{L2P~\citep{wang2022learning}}
& \multicolumn{2}{c|}{DualPrompt~\citep{wang2022dualprompt}}
& \multicolumn{2}{c|}{CODA-Prompt~\citep{Smith2023CODAprompt}}
& \multicolumn{2}{c}{APT~\citep{APT_2025_ICCV}} \\
& Avg. Acc ($\uparrow$) & Forgetting ($\downarrow$)
& Avg. Acc ($\uparrow$) & Forgetting ($\downarrow$)
& Avg. Acc ($\uparrow$) & Forgetting ($\downarrow$)
& Avg. Acc ($\uparrow$) & Forgetting ($\downarrow$) \\
\midrule
Baseline (w/o Distillation)
& 73.94 $\pm$ 0.22 & 4.41 $\pm$ 0.18
& 73.18 $\pm$ 0.33 & 3.45 $\pm$ 0.32
& 76.42 $\pm$ 0.17 & 4.31 $\pm$ 0.18
& 78.67 $\pm$ 0.28 & 4.06 $\pm$ 0.29 \\
\midrule
CDL-KD~\citep{hinton2015distilling}
& 74.12 $\pm$ 0.42 & 4.60 $\pm$ 0.55
& 73.90 $\pm$ 0.14 & \textbf{3.31 $\pm$ 0.04}
& 76.99 $\pm$ 0.02 & 3.81 $\pm$ 0.06
& 79.21 $\pm$ 0.10 & 3.74 $\pm$ 0.19 \\
CDL-DKD~\citep{zhao2022decoupled}
& 74.58 $\pm$ 0.01 & 4.69 $\pm$ 0.06
& 75.24 $\pm$ 0.33 & 4.15 $\pm$ 0.23
& 76.70 $\pm$ 0.17 & 4.84 $\pm$ 0.12
& 78.85 $\pm$ 0.21 & \textbf{2.69 $\pm$ 0.31} \\
CDL-FitNets~\citep{romero2014fitnets}
& 70.39 $\pm$ 0.23 & 5.84 $\pm$ 0.06
& 71.23 $\pm$ 0.04 & 5.71 $\pm$ 0.55
& 74.55 $\pm$ 0.14 & 6.81 $\pm$ 0.15
& 78.88 $\pm$ 0.11 & 4.38 $\pm$ 0.18 \\
CDL-ReviewKD~\citep{chen2021ReviewKD}
& 72.17 $\pm$ 0.26 & 6.11 $\pm$ 0.08
& 72.19 $\pm$ 0.01 & 5.72 $\pm$ 0.20
& 75.72 $\pm$ 0.27 & 4.14 $\pm$ 0.05
& 78.79 $\pm$ 0.23 & 4.15 $\pm$ 0.07 \\
CDL-DeiT~\citep{DeiT2020}
& 73.99 $\pm$ 0.01 & 5.09 $\pm$ 0.02
& 76.03 $\pm$ 0.03 & 3.90 $\pm$ 0.01
& 77.83 $\pm$ 0.55 & 4.51 $\pm$ 0.03
& 79.43 $\pm$ 0.11 & 3.61 $\pm$ 0.16 \\
\midrule
\textbf{D-CDL (ours)}
& \textbf{76.91 $\pm$ 0.40} & \textbf{3.15 $\pm$ 0.39}
& \textbf{76.06 $\pm$ 0.12} & 3.77 $\pm$ 0.38
& \textbf{78.62 $\pm$ 0.57} & \textbf{3.46 $\pm$ 0.53}
& \textbf{79.83 $\pm$ 0.19} & 3.87 $\pm$ 0.19 \\
\midrule
\bottomrule
\end{tabular}
}
\end{table}

% Table #######################
\begin{table}[!h]
\centering
\caption{
Continual distillation results on the CIFAR-100 dataset with ViT-Small students (ViT-Base teachers) under different prompt-based continual learning frameworks.
}
\label{tab:cifar100_small_all_methods}
\renewcommand{\arraystretch}{1.0}
\resizebox{\textwidth}{!}{
\begin{tabular}{l|cc|cc|cc|cc}
\toprule
\multirow{2}{*}{Methods}
& \multicolumn{2}{c|}{L2P~\citep{wang2022learning}}
& \multicolumn{2}{c|}{DualPrompt~\citep{wang2022dualprompt}}
& \multicolumn{2}{c|}{CODA-Prompt~\citep{Smith2023CODAprompt}}
& \multicolumn{2}{c}{APT~\citep{APT_2025_ICCV}} \\
& Avg. Acc ($\uparrow$) & Forgetting ($\downarrow$)
& Avg. Acc ($\uparrow$) & Forgetting ($\downarrow$)
& Avg. Acc ($\uparrow$) & Forgetting ($\downarrow$)
& Avg. Acc ($\uparrow$) & Forgetting ($\downarrow$) \\
\midrule
Baseline (w/o Distillation)
& 77.71 $\pm$ 0.49 & 7.12 $\pm$ 0.33
& 79.85 $\pm$ 0.57 & 6.12 $\pm$ 0.32
& 82.18 $\pm$ 0.20 & 6.48 $\pm$ 0.48
& 83.30 $\pm$ 0.15 & 4.35 $\pm$ 0.09 \\
\midrule
CDL-KD~\citep{hinton2015distilling}
& 79.64 $\pm$ 0.04 & 6.35 $\pm$ 0.02
& 80.16 $\pm$ 0.54 & 5.76 $\pm$ 0.18
& 83.03 $\pm$ 0.39 & 7.24 $\pm$ 0.30
& 84.27 $\pm$ 0.05 & 4.67 $\pm$ 0.13 \\
CDL-DKD~\citep{zhao2022decoupled}
& 78.21 $\pm$ 0.12 & 9.13 $\pm$ 0.07
& 80.44 $\pm$ 0.46 & 6.96 $\pm$ 0.44
& 82.27 $\pm$ 0.20 & 7.81 $\pm$ 0.14
& 83.51 $\pm$ 0.30 & 4.29 $\pm$ 0.03 \\
CDL-FitNets~\citep{romero2014fitnets}
& 79.56 $\pm$ 0.39 & 5.89 $\pm$ 0.36
& 80.70 $\pm$ 0.17 & 5.73 $\pm$ 0.21
& 81.83 $\pm$ 0.05 & 8.83 $\pm$ 0.48
& 83.95 $\pm$ 0.22 & 4.64 $\pm$ 0.27 \\
CDL-ReviewKD~\citep{chen2021ReviewKD}
& 78.50 $\pm$ 0.39 & 8.04 $\pm$ 0.75
& 80.33 $\pm$ 0.23 & 5.86 $\pm$ 0.53
& 82.20 $\pm$ 0.41 & 7.54 $\pm$ 0.03
& 84.08 $\pm$ 0.01 & 4.82 $\pm$ 0.08 \\
CDL-DeiT~\citep{DeiT2020}
& 79.56 $\pm$ 0.07 & 6.71 $\pm$ 0.16
& 80.64 $\pm$ 0.32 & 5.67 $\pm$ 0.59
& 83.79 $\pm$ 0.15 & 6.58 $\pm$ 0.06
& 83.81 $\pm$ 0.01 & \textbf{3.44 $\pm$ 0.34} \\
\midrule
\textbf{D-CDL (ours)}
& \textbf{81.79 $\pm$ 0.66} & \textbf{4.31 $\pm$ 0.27}
& \textbf{81.78 $\pm$ 0.17} & \textbf{3.63 $\pm$ 0.03}
& \textbf{84.31 $\pm$ 0.01} & \textbf{5.63 $\pm$ 0.02}
& \textbf{84.28 $\pm$ 0.11} & 3.89 $\pm$ 0.09 \\
\midrule
\bottomrule
\end{tabular}
}
\end{table}

% Table #######################
\begin{table}[!h]
\centering
\caption{
Continual distillation results on the CIFAR-100 dataset with ViT-Base students (ViT-Large teachers) under different prompt-based continual learning frameworks.
}
\label{tab:cifar100_base_all_methods}
\renewcommand{\arraystretch}{1.0}
\resizebox{\textwidth}{!}{
\begin{tabular}{l|cc|cc|cc|cc}
\toprule
\multirow{2}{*}{Methods}
& \multicolumn{2}{c|}{L2P~\citep{wang2022learning}}
& \multicolumn{2}{c|}{DualPrompt~\citep{wang2022dualprompt}}
& \multicolumn{2}{c|}{CODA-Prompt~\citep{Smith2023CODAprompt}}
& \multicolumn{2}{c}{APT~\citep{APT_2025_ICCV}} \\
& Avg. Acc ($\uparrow$) & Forgetting ($\downarrow$)
& Avg. Acc ($\uparrow$) & Forgetting ($\downarrow$)
& Avg. Acc ($\uparrow$) & Forgetting ($\downarrow$)
& Avg. Acc ($\uparrow$) & Forgetting ($\downarrow$) \\
\midrule
Baseline (w/o Distillation)
& 83.02 $\pm$ 0.47 & 6.06 $\pm$ 0.47
& 84.66 $\pm$ 0.87 & 5.91 $\pm$ 0.34
& 86.16 $\pm$ 0.17 & 5.63 $\pm$ 0.25
& 88.26 $\pm$ 0.09 & 3.31 $\pm$ 0.33 \\
\midrule
CDL-KD~\citep{hinton2015distilling}
& 85.00 $\pm$ 0.34 & \textbf{4.48 $\pm$ 0.51}
& 84.67 $\pm$ 0.53 & \textbf{4.52 $\pm$ 0.55}
& 86.27 $\pm$ 0.05 & 5.45 $\pm$ 0.06
& 88.79 $\pm$ 0.26 & 2.85 $\pm$ 0.21 \\
CDL-DKD~\citep{zhao2022decoupled}
& 83.29 $\pm$ 0.24 & 4.99 $\pm$ 0.17
& 84.93 $\pm$ 0.16 & 4.95 $\pm$ 0.12
& 85.42 $\pm$ 0.31 & 6.55 $\pm$ 0.16
& 88.69 $\pm$ 0.03 & \textbf{1.79 $\pm$ 0.02} \\
CDL-FitNets~\citep{romero2014fitnets}
& 83.60 $\pm$ 0.02 & 5.21 $\pm$ 0.71
& 83.12 $\pm$ 0.86 & 8.33 $\pm$ 1.36
& 85.95 $\pm$ 0.25 & 6.56 $\pm$ 0.02
& 88.00 $\pm$ 0.27 & 3.35 $\pm$ 0.18 \\
CDL-ReviewKD~\citep{chen2021ReviewKD}
& 83.12 $\pm$ 0.65 & 7.97 $\pm$ 0.27
& 84.11 $\pm$ 0.92 & 5.19 $\pm$ 0.33
& 86.21 $\pm$ 0.61 & 5.64 $\pm$ 0.40
& 88.60 $\pm$ 0.04 & 3.00 $\pm$ 0.17 \\
CDL-DeiT~\citep{DeiT2020}
& 84.21 $\pm$ 0.71 & 6.06 $\pm$ 0.93
& 85.73 $\pm$ 0.27 & 5.05 $\pm$ 0.43
& 86.78 $\pm$ 0.15 & 5.43 $\pm$ 0.25
& 88.72 $\pm$ 0.13 & 3.15 $\pm$ 0.40 \\
\midrule
\textbf{D-CDL (ours)}
& \textbf{86.56 $\pm$ 0.22} & 4.97 $\pm$ 0.07
& \textbf{86.92 $\pm$ 0.24} & 4.77 $\pm$ 0.58
& \textbf{87.13 $\pm$ 0.09} & \textbf{5.30 $\pm$ 0.06}
& \textbf{89.41 $\pm$ 0.04} & 2.74 $\pm$ 0.09 \\
\midrule
\bottomrule
\end{tabular}
}
\end{table}

\subsection{Additional Results on CUB-200}
\label{appendix:cub200}
To further evaluate the generalization ability of D-CDL beyond the two main benchmarks, we additionally conduct experiments on CUB-200~\citep{wah2011caltech}, a fine-grained visual recognition benchmark comprising 200 bird species. Following the standard class-incremental setting, we divide the dataset into 10 tasks with 20 classes per task and adopt the ViT-Base $\rightarrow$ ViT-Small distillation setting. Unless otherwise specified, all other experimental settings are kept consistent with those used for the main experiments.

\begin{table*}[t]

\centering
\caption{Additional results on CUB-200 under the ViT-Base $\rightarrow$ ViT-Small distillation setting with 10 tasks and 20 classes per task.}
\label{tab:cub200_results}
{
\resizebox{\linewidth}{!}{
\begin{tabular}{l cc cc cc cc}
\toprule
\multirow{2}{*}{Methods}
& \multicolumn{2}{c}{L2P~\citep{wang2022learning}}
& \multicolumn{2}{c}{DualPrompt~\citep{wang2022dualprompt}}
& \multicolumn{2}{c}{CODA-Prompt~\citep{Smith2023CODAprompt}}
& \multicolumn{2}{c}{APT~\citep{APT_2025_ICCV}} \\
\cmidrule(lr){2-3}
\cmidrule(lr){4-5}
\cmidrule(lr){6-7}
\cmidrule(lr){8-9}
& Avg. Acc ($\uparrow$) & Forgetting ($\downarrow$)
& Avg. Acc ($\uparrow$) & Forgetting ($\downarrow$)
& Avg. Acc ($\uparrow$) & Forgetting ($\downarrow$)
& Avg. Acc ($\uparrow$) & Forgetting ($\downarrow$) \\
\midrule
Baseline (w/o Distillation)
& 71.52 & 7.33
& 72.61 & 6.90
& 74.15 & 7.09
& 76.36 & 5.39 \\
CDL-KD~\citep{hinton2015distilling}
& 72.56 & 6.59
& 73.79 & 6.68
& 75.48 & 6.84
& 75.74 & 5.72 \\
CDL-FitNets~\citep{romero2014fitnets}
& 72.69 & 6.62
& 73.79 & 6.66
& 75.55 & 6.85
& 75.73 & 5.89 \\
\midrule
\textbf{D-CDL (ours)}
& \textbf{74.72} & \textbf{4.67}
& \textbf{75.55} & \textbf{4.21}
& \textbf{77.15} & \textbf{5.78}
& \textbf{77.37} & \textbf{4.21} \\
\bottomrule
\end{tabular}
}
}
\end{table*}

As shown in Table~\ref{tab:cub200_results}, D-CDL achieves the highest average accuracy and the lowest forgetting across all prompt-based continual learning methods. These results further demonstrate the effectiveness of D-CDL beyond the two main benchmarks and its generalization to fine-grained visual recognition.

\subsection{Mechanistic Analysis of Decoupled Distillation}
\label{appendix:mechanistic_analysis}

To better understand what is decoupled by D-CDL, we provide two additional analyses: branch-level output alignment and cross-task prompt stability. Both analyses are conducted on Split ImageNet-R under the CODA-Prompt framework with the ViT-Base $\rightarrow$ ViT-Small distillation setting.

\textbf{Two-branch output alignment.}
After training all 10 tasks, we randomly sample 1,000 test images from all 10 tasks and compare the outputs of the two student branches with the teacher output. Specifically, we compare the logits from the original student CLS head and the student KD head against the teacher logits. We report the KL divergence to the teacher distribution, cosine similarity between logits, and top-1 agreement with the teacher prediction.

\begin{table}[t]

\centering
\caption{Two-branch output alignment analysis on Split ImageNet-R. The model is evaluated after training all 10 tasks, and 1,000 test images are randomly sampled from all 10 tasks.}
\label{tab:branch_alignment}
{
\resizebox{0.78\linewidth}{!}{
\begin{tabular}{l c c c}
\toprule
Branch & KL to Teacher ($\downarrow$) & Logit Cosine ($\uparrow$) & Teacher Agree. (\%) ($\uparrow$) \\
\midrule
Student CLS head & 2.234 & 0.653 & 67.7 \\
Student KD head  & \textbf{2.185} & \textbf{0.681} & \textbf{68.4} \\
\bottomrule
\end{tabular}
}
}
\end{table}

As shown in Table~\ref{tab:branch_alignment}, the student KD head is more closely aligned with the teacher output than the original student CLS head, with lower KL divergence, higher logit cosine similarity, and higher teacher agreement. This supports our decoupled design: the KD branch is more teacher-aligned, while the original CLS branch remains responsible for task-specific classification.

\textbf{Cross-task prompt stability.}
We further analyze the stability of prompt parameters across consecutive task checkpoints. Specifically, after each task is trained, we compare the saved prompt parameters with those from the previous task. We report the average cosine similarity and relative $\ell_2$ drift across all consecutive task pairs. The relative $\ell_2$ drift is computed as the ratio between the parameter change and the norm of the previous-task parameters.

\begin{table}[t]

\centering
\caption{Prompt stability across consecutive task checkpoints on Split ImageNet-R. Higher cosine similarity and lower relative $\ell_2$ drift indicate smoother cross-task evolution.}
\label{tab:prompt_stability}
{
\resizebox{0.5\linewidth}{!}{
\begin{tabular}{l c c}
\toprule
Prompt Type & Cosine Similarity ($\uparrow$) & Relative $\ell_2$ Drift ($\downarrow$) \\
\midrule
KD-prompts & \textbf{0.995} & \textbf{0.098} \\
CL prompts & 0.928 & 0.390 \\
\bottomrule
\end{tabular}
}
}
\end{table}

As shown in Table~\ref{tab:prompt_stability}, KD-prompts are substantially more stable than the original CL prompts across consecutive task checkpoints. This is consistent with our design: KD-prompts are propagated across tasks to serve as a persistent carrier for teacher knowledge, while the original CL prompts are more task-adaptive and therefore change more significantly across tasks.

\subsection{Multiple Layers \& KD-Prompts Length}
\label{appendix:Layers_Length}
\begin{figure}
  \centering
  \includegraphics[width=0.7\linewidth]{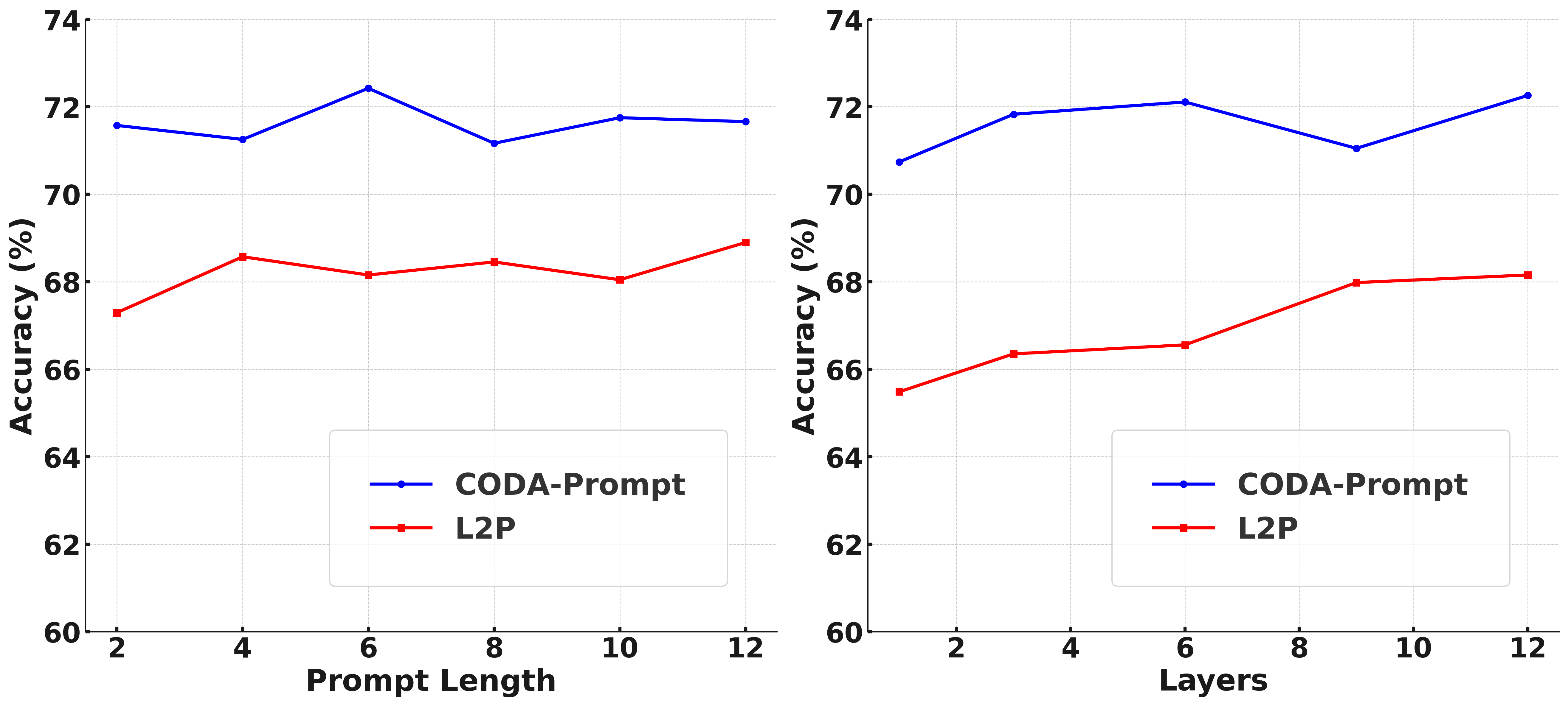}
\caption{Left: Effect of the KD-prompt length on accuracy. Right: Effect of the number of inserted KD-prompt layers on accuracy. Results are reported on Split ImageNet-R under the ViT-Base $\rightarrow$ ViT-Small distillation setting.}
  \label{fig:Layers_acc}
  % \vspace{-4mm}
\end{figure}

We conduct experiments under the CODA-Prompt~\citep{Smith2023CODAprompt} and L2P~\citep{wang2022learning} frameworks with 10 tasks. In our model, a KD-prompt is denoted by $\mathbf{P}^{\mathrm{d}} \in \mathbb{R}^{L_p \times D}$. Fig.~\ref{fig:Layers_acc} (left) shows the effect of the KD-prompt length $L_p$ on accuracy when the KD-prompts are inserted into all 12 transformer blocks. Fig.~\ref{fig:Layers_acc} (right) illustrates the effect of the number of inserted layers $n$ on accuracy with $L_p=6$. The results show that the KD-prompt length has limited influence on performance, while the number of inserted layers is positively correlated with accuracy, especially for L2P. Therefore, we choose $n=12$, i.e., inserting KD-prompts into all transformer blocks, for the best performance.

% ============================================================
% New appendix sections added during rebuttal
% ============================================================

\subsection{More Ablation Studies}
\label{appendix:more_ablation}

\textbf{Parameter-matched and two-head baseline.}
To disentangle the effect of the proposed decoupled prompting mechanism from both additional trainable parameters and two-head ensembling, we construct a parameter-matched baseline. Specifically, for each prompt-based continual learning method, we introduce the same number of additional prompt parameters as D-CDL. Following the original prompt insertion strategy of each method, for CODA-Prompt, we add extra prompts to the first five transformer layers with the same prompt length used by D-CDL, while for L2P, we add extra prompts only to the first transformer layer. For a fair comparison, D-CDL is also restricted to the same prompt insertion layers in each setting. In addition, the baseline uses the same two-head classifier design as D-CDL, but both classifiers are attached to the original CLS-token representation. Therefore, this baseline does not introduce the KD token or the decoupled KD-token/KD-prompt pathway. One classifier is supervised by ground-truth labels, while the other classifier is supervised by teacher logits through KD (the same as D-CDL). During inference, the final prediction is obtained by combining the outputs of the two classifiers, following the same protocol as D-CDL.

\begin{table}[t]

\centering
\caption{Parameter-matched and two-head baseline on Split ImageNet-R. The baseline uses the same extra prompt parameters and the same two-head inference protocol as D-CDL, while removing the proposed decoupled KD-token/KD-prompt pathway.}
\label{tab:param_matched_two_head}
{
\resizebox{0.85\linewidth}{!}{
\begin{tabular}{l l c c}
\toprule
Framework & Method & Avg. Acc ($\uparrow$) & Forgetting ($\downarrow$) \\
\midrule
L2P
& KD + matched extra prompts + two-head classifier
& 65.47 & 6.43 \\
L2P
& \textbf{D-CDL (ours)}
& \textbf{66.41} & \textbf{3.21} \\
\midrule
CODA-Prompt
& KD + matched extra prompts + two-head classifier
& 70.46 & 7.21 \\
CODA-Prompt
& \textbf{D-CDL (ours)}
& \textbf{71.86} & \textbf{5.64} \\
\bottomrule
\end{tabular}
}
}
\end{table}

As shown in Table~\ref{tab:param_matched_two_head}, even when the baseline is given the same number of additional trainable prompt parameters and the same two-head inference protocol, D-CDL still achieves higher accuracy and lower forgetting. Since both heads in the baseline are connected to the same CLS-token branch, this comparison controls for the effect of two-head ensembling without introducing decoupled distillation. These results suggest that the gains of D-CDL cannot be solely explained by additional prompt parameters or ensemble effects from using two classifiers. Instead, they support the effectiveness of the proposed decoupled prompting design, where task-specific adaptation and teacher knowledge transfer are separated through the KD-token/KD-prompt pathway.

\textbf{KD-prompt attachment and propagation.}
We further conduct an ablation study to examine the effect of where KD-prompts are attached and whether they are propagated across tasks. This experiment is conducted on Split ImageNet-R under the CODA-Prompt framework with the ViT-Base $\rightarrow$ ViT-Small distillation setting. For fairness, all variants use the same number and length of KD-prompts inserted into all 12 transformer layers of the ViT-Small student backbone, where the lengths of $\mathbf{P}_K^d$ and $\mathbf{P}_V^d$ in each block are both set to $3$.

\begin{table}[t]

\centering
\caption{Ablation on KD-prompt attachment and KD-prompt propagation. All variants are evaluated on Split ImageNet-R under the CODA-Prompt framework with the ViT-Base $\rightarrow$ ViT-Small distillation setting.}
\label{tab:kd_prompt_attachment_propagation}
{
\resizebox{0.85\linewidth}{!}{
\begin{tabular}{l c c c}
\toprule
Method & KD-Prompt Attachment & KD-Prompt Propagation & Avg. Acc ($\uparrow$) \\
\midrule
KD-prompts attached to CLS token
& CLS token & $\checkmark$ & 71.29 \\
D-CDL w/o KD-prompt propagation
& KD token & $\varnothing$ & 71.02 \\
\textbf{D-CDL (ours)}
& \textbf{KD token} & $\checkmark$ & \textbf{71.92} \\
\bottomrule
\end{tabular}
}
}
\end{table}

All variants use the same number of additional KD-prompt parameters and the same two-head prediction protocol. The first variant attaches KD-prompts to the original CLS-token branch together with the CL prompts, while still allowing KD-prompts to be propagated across tasks. The second variant keeps the decoupled KD-token branch but disables KD-prompt propagation by re-initializing KD-prompts at the beginning of each task. The full D-CDL model uses both the decoupled KD-token branch and cross-task KD-prompt propagation. Since all variants use the same amount of extra prompts and combine two classifier outputs during inference, this comparison controls for both additional prompt capacity and possible ensemble effects between two classifiers.

As shown in Table~\ref{tab:kd_prompt_attachment_propagation}, the full D-CDL setting achieves the best performance among the compared variants. These results support our decoupled prompting design: KD-prompts are most effective when they are both separated from the task-adaptive CLS branch and propagated across tasks as a persistent carrier for distillation knowledge.

\subsection{Logit Distillation Details}
\label{appendix:Logit_KD}

Logit distillation is one of the most classic forms of knowledge distillation. It aims to have a smaller student model learn the logits output of a larger or more accurate teacher model. In our CDL setup, we conducted experiments using two types of logit knowledge distillation methods: normal knowledge distillation (KD)~\citep{hinton2015distilling} and Decoupled Knowledge Distillation (DKD)~\citep{zhao2022decoupled}.

\label{sec:Logits-Distillation}

% This method is closely related to the traditional "hard target." In model training, the loss function is defined to bring the model’s predicted value closer to the true value, which represents the hard target.

\textbf{Normal Knowledge Distillation (KD):} This approach was initially proposed by ~\cite{hinton2015distilling}, where distillation transfers the knowledge from a large, complex teacher model to a smaller student model, helping the latter to approximate the teacher model in terms of performance and accuracy. To achieve this, Hinton et al. designed a method where the logit output of the teacher model serves as ``soft labels'', guiding the student model’s training. After passing through the softmax layer, the output values provide probabilities for each class.
\begin{equation}
p_{i}= \frac{\exp(z_i / \tau)}{\sum_j \exp(z_j / \tau)},
\label{soft_target}
\end{equation}
where $z_i$ is the logit, and $p_i$ is the predicted probability for class $i$. The soft target is the class probability distribution of teacher model. In the distillation process, a temperature parameter $\tau$, is introduced to smooth the output distribution of the model, making it easier for the student model to learn the subtle differences between classes. 

The class tokens of the teacher model and the student model processed by the pre-trained ViT backbones are connected to the teacher classifier and student classifier to output their logits, respectively. The loss function of the student model for logit distillation is defined as:
\begin{align}
\mathcal{L}_{\text{S}} &= (1 - \alpha) \mathcal{L}(g_{\phi}(f_{b}(\mathbf{x})), y) + \alpha  \mathcal{L}_{\text{KD}} + \lambda \mathcal{L}_{\text{pool}}, \label{Loss:logit} \\
 \mathcal{L}_{\text{KD}} &= \tau^2 \sum_{i} p_{i}^{\mathcal{T}} \log\left(\frac{p_{i}^{\mathcal{T}}}{p_{i}^{\mathcal{S}}}\right),
\label{Loss:KD}
\end{align}
where $ \mathcal{L}(g_{\phi}(f_{b}(\mathbf{x})), y) $ represents the cross-entropy classification loss used for learning with true label $y$, $g_{\phi}$ is the student classifier, and $f_{b}$ denotes the pre-trained ViT backbone and we only use the final class token into the classifier.  $\mathcal{L}_{\text{KD}} $ represents the knowledge distillation loss, i.e., the KL divergence between the teacher's probability distribution $p_{i}^{\mathcal{T}}$ and the student's probability distribution $p_{i}^{\mathcal{S}}$. The \( \mathcal{L}_{\text{pool}} \) is a loss function specific to the prompt pool. Different prompt-based continual learning methods have their respective prompt pool loss functions. For example, please refer to the loss functions in~\eqref{eq:l2p_loss_short} and~\eqref{eq:coda_loss_short} for L2P and CODA-Prompt. \( \alpha \) and $\lambda$ are hyperparameters used to balance the weights of loss components.

\textbf{Decoupled Knowledge Distillation (DKD):} Unlike traditional knowledge distillation, which uses a unified KL divergence to measure the difference between the outputs of the student model and the teacher model, DKD separates the logit distillation loss into target class and non-target class components. DKD considers that the target and non-target classes contain different information and should be handled separately during training. By decoupling these components, DKD allows the student model to better capture the confidence on the target class while learning the distribution of the non-target classes. The student model loss function is formulated as follows:
\begin{align}
\mathcal{L}_{\text{S}}
&= (1-\alpha)
\mathcal{L}\left(g_{\phi}(f_b(\mathbf{x})),y\right)
\nonumber\\
&\quad
+\alpha\left(
\mathcal{L}_{\text{TCKD}}
+\mathcal{L}_{\text{NCKD}}
\right)
+\lambda\mathcal{L}_{\text{pool}},
\\
\mathcal{L}_{\text{TCKD}}
&=
p_t^{\mathcal{T}}
\log\left(
\frac{p_t^{\mathcal{T}}}{p_t^{\mathcal{S}}}
\right)
+
p_{\setminus t}^{\mathcal{T}}
\log\left(
\frac{p_{\setminus t}^{\mathcal{T}}}
     {p_{\setminus t}^{\mathcal{S}}}
\right),
\\
\mathcal{L}_{\text{NCKD}}
&=
\sum_{i\neq t}
\hat{p}_i^{\mathcal{T}}
\log\left(
\frac{\hat{p}_i^{\mathcal{T}}}
{\hat{p}_i^{\mathcal{S}}}
\right).
\end{align}
where $\mathcal{L}_{\text{TCKD}}$ is the target class distillation loss, focused on aligning the student’s confidence with the teacher’s for the correct class. $\mathcal{L}_{\text{NCKD}}$ is the non-target class distillation loss, focused on matching the teacher and student model distributions for the incorrect classes. $[p_t^{\mathcal{T}}, p_{\setminus t}^{\mathcal{T}}]$ represents the binary probabilities of the target class $p_t^{T}$ and all the other non-target classes $p_{\setminus t}^{\mathcal{T}}$ in teacher model, which can be calculated by~\eqref{soft_target}. $[p_t^{\mathcal{S}}, p_{\setminus t}^{\mathcal{S}}]$ represents the binary probabilities in student model. Meanwhile, $\hat{p}_i^{\mathcal{T}}$ and $\hat{p}_i^{\mathcal{S}}$ are probability distributions among the non-target classes (without considering the $t$th class). Each element is calculated by:
\begin{equation}
\hat{p}_i = \frac{\exp(z_i/ \tau)}{\sum\limits_{j \neq t}\exp(z_j/ \tau)}.
\end{equation}

\vspace{-4mm}
\subsection{Feature Distillation Details}
\label{appendix:Feature_KD}

Feature distillation focuses on transferring the intermediate representations from a teacher model to a student model. It can leverage richer, layer-wise information within the teacher model to guide the student model’s learning. The student model can benefit from an understanding of feature relationships. In the CDL model, the backbone ViT consists of multiple blocks, and we use the internal tokens outputted by each block as features. This allows the internal tokens of the student model to learn the information in the internal tokens of the teacher model. It is worth noting that during knowledge distillation, the backbone of the student model remains frozen while processing the internal tokens, which is a characteristic of prompt-based continual learning methods. In this paper, we build two methods based on the handling of internal tokens: FitNets~\citep{romero2014fitnets} and Review Knowledge Distillation (ReviewKD)~\citep{chen2021ReviewKD}.

\textbf{FitNets:} This method is one of the most classic feature distillation methods, aimed at adding a distillation loss to the intermediate layers. It uses the intermediate feature representations of the teacher model as hints to guide the student model's learning. Here, we select the output of the last block of the teacher model as the hint. Similarly, the student model selects the output of the corresponding block for learning, constructing the feature distillation loss. Finally, the total loss function for the student model is
\begin{align}
\mathcal{L}_{\text{S}} &=  \mathcal{L}(g_{\phi}(f_{b}(\mathbf{x})), y) + \alpha  \mathcal{L}_{\text{hint}} + \lambda \mathcal{L}_{\text{pool}}, \\
\mathcal{L}_{\text{hint}} &= \left\| f_{b-1}^{\mathcal{T}}(\mathbf{x}) - F_{M}(f_{b-1}^{\mathcal{S}}(\mathbf{x})) \right\|^2,
\end{align}
where $\mathcal{L}_{\text{hint}}$ is the feature distillation loss, calculated using the Mean Squared Error (MSE). $\alpha$ and $\lambda$ are used to balance the weights of loss components. $f_{b-1}$ indicates the feature output (internal tokens) after the last block of the ViT model. The student feature is transformed into the same size as the teacher feature with the mapping layer $F_{M}$, which is simply a fully-connected layer in the network.

\textbf{ReviewKD:} It innovates by ``reviewing'' multiple hidden layers from both the teacher and student models, offering a more comprehensive approach to capturing hierarchical features across the entire model. It reviews the multiple layers utilizing the concept of residual learning. For instance, the feature from $n_{th}$ block of the student is aggregated with the feature from $(n-1)_{th}$ block of the student to mimic the feature from $(n-1)_{th}$ block of the teacher. The total loss function of the student in ReviewKD is
\begin{align}
&\mathcal{L}_{\text{S}} =  \mathcal{L}(g_{\phi}(f_{b}(\mathbf{x})), y) + \alpha  \mathcal{L}_{\text{RKD}} + \lambda \mathcal{L}_{\text{pool}}, \\
& \mathcal{L}_{\text{RKD}} = \mathcal{D}(\mathbf{F}^\mathcal{S}_n, \mathbf{F}^\mathcal{T}_n) + \sum_{j=1}^{n-1} \mathcal{D} \left( \mathcal{U}(\mathbf{F}_j^\mathcal{S}, \mathbf{F}_{j+1}^\mathcal{S}), \mathbf{F}^\mathcal{T}_j \right),
\label{reviewKD}
\end{align}
where $\mathcal{L}_{\text{RKD}}$ is the reviewKD loss, and $\mathcal{D}$ is L2 distance between the student features and teacher features. All student features in the equations have passed through the mapping layer \( F_M \) to make them the same dimension as the teacher features. $\mathbf{F}_j^\mathcal{S}$ and $\mathbf{F}_j^\mathcal{T}$ are the features output by the student model and teacher model, respectively, after passing through \( j \) blocks. $\mathbf{F}_{j+1}^\mathcal{S}$ represents the fused student features at the $(j+1)_{th}$ block.  $\mathcal{U}$ is a module used to fuse features, which performs a weighted combination of the two input features. \( n \) is the total number of blocks in the ViT backbone.  Therefore, in the student model, the fused features obtained at each block are passed to the next higher block to form new feature fusion.

\end{document}

%% file: math_commands.tex
\usepackage{amsmath,amsfonts,bm}

\def\eqref#1{equation~\ref{#1}}
\def\1{\bm{1}}

\DeclareMathAlphabet{\mathsfit}{\encodingdefault}{\sfdefault}{m}{sl}
\SetMathAlphabet{\mathsfit}{bold}{\encodingdefault}{\sfdefault}{bx}{n}

%% file: collas2026_conference.bib
@String(CVPR= {IEEE Conf. Comput. Vis. Pattern Recog.})

@String(ICCV= {Int. Conf. Comput. Vis.})

@String(ECCV= {Eur. Conf. Comput. Vis.})

@String(ICPR = {Int. Conf. Pattern Recog.})

@String(ICLR = {Int. Conf. Learn. Represent.})

@String(AAAI = {AAAI})

@String(CVPR  = {CVPR})

@String(ICCV  = {ICCV})

@String(ECCV  = {ECCV})

@String(ICPR  = {ICPR})

@String(ICLR  = {ICLR})

@inproceedings{rebuffi2017icarl,
  title={icarl: Incremental classifier and representation learning},
  author={Rebuffi, Sylvestre-Alvise and Kolesnikov, Alexander and Sperl, Georg and Lampert, Christoph H},
  booktitle={Proceedings of the IEEE conference on Computer Vision and Pattern Recognition},
  pages={2001--2010},
  year={2017}
}

@inproceedings{wang2022learning,
  title={Learning to prompt for continual learning},
  author={Wang, Zifeng and Zhang, Zizhao and Lee, Chen-Yu and Zhang, Han and Sun, Ruoxi and Ren, Xiaoqi and Su, Guolong and Perot, Vincent and Dy, Jennifer and Pfister, Tomas},
  booktitle={Proceedings of the IEEE/CVF Conference on Computer Vision and Pattern Recognition},
  pages={139--149},
  year={2022}
}

@inproceedings{wang2022dualprompt,
  title={Dualprompt: Complementary prompting for rehearsal-free continual learning},
  author={Wang, Zifeng and Zhang, Zizhao and Ebrahimi, Sayna and Sun, Ruoxi and Zhang, Han and Lee, Chen-Yu and Ren, Xiaoqi and Su, Guolong and Perot, Vincent and Dy, Jennifer and others},
  booktitle={European Conference on Computer Vision},
  pages={631--648},
  year={2022},
  organization={Springer}
}

@InProceedings{Smith2023CODAprompt,
    author = {Smith, James Seale and Karlinsky, Leonid and Gutta, Vyshnavi and Cascante-Bonilla, Paola and Kim, Donghyun and Arbelle, Assaf and Panda, Rameswar and Feris, Rogerio and Kira, Zsolt},
    title = {CODA-Prompt: COntinual Decomposed Attention-Based Prompting for Rehearsal-Free Continual Learning},
    booktitle = {Proceedings of the IEEE/CVF Conference on Computer Vision and Pattern Recognition (CVPR)},
    month = {June},
    year = {2023},
    pages = {11909-11919}
}

@inproceedings{isele2018selective,
  title={Selective Experience Replay for Lifelong Learning},
  author={Isele, David and Cosgun, Akansel},
  booktitle={Proceedings of the AAAI Conference on Artificial Intelligence},
  pages={3302--3309},
  year={2018}
}

@inproceedings{castro2018end,
  title={End-to-end incremental learning},
  author={Castro, Francisco M and Mar{\'\i}n-Jim{\'e}nez, Manuel J and Guil, Nicol{\'a}s and Schmid, Cordelia and Alahari, Karteek},
  booktitle={Proceedings of the European Conference on Computer Vision (ECCV)},
  pages={233--248},
  year={2018},
  organization={Springer}
}

@inproceedings{chaudhry2019tinyMemory,
  title={Continual Learning with Tiny Episodic Memories},
  author={Chaudhry, Arslan and Ranzato, Marc'Aurelio and Rohrbach, Marcus and Elhoseiny, Mohamed},
  booktitle={Proceedings of the 33rd Conference on Neural Information Processing Systems (NeurIPS)},
  year={2019}
}

@inproceedings{chaudhry2019GEMA,
  title={Efficient Lifelong Learning with A-GEM},
  author={Chaudhry, Arslan and Rohrbach, Marcus and Elhoseiny, Mohamed and Ajanthan, Thalaiyasingam and Dokania, Puneet K and Torr, Philip HS and Ranzato, Marc'Aurelio},
  booktitle={International Conference on Learning Representations (ICLR)},
  year={2019}
}

@Article{kirkpatrick2017overcoming_EWC,
  title={Overcoming catastrophic forgetting in neural networks},
  author={Kirkpatrick, James and Pascanu, Razvan and Rabinowitz, Neil and Veness, Joel and Desjardins, Guillaume and Rusu, Andrei A and Milan, Kieran and Quan, John and Ramalho, Tiago and Grabska-Barwinska, Agnieszka and others},
  journal={Proceedings of the National Academy of Sciences},
  volume={114},
  number={13},
  pages={3521--3526},
  year={2017},
  publisher={National Acad Sciences}
}

@article{lopezpaz2017gradient,
  title={Gradient Episodic Memory for Continual Learning},
  author={Lopez-Paz, David and Ranzato, Marc'Aurelio},
  journal={arXiv preprint arXiv:1706.08840},
  year={2017}
}

@misc{hinton2015distilling,
  title={Distilling the Knowledge in a Neural Network},
  author={Hinton, Geoffrey and Vinyals, Oriol and Dean, Jeff},
  archivePrefix={arXiv},
  eprint={1503.02531},
  primaryClass={stat.ML},
  year={2015}
}

@inproceedings{liang2023less,
  title={Less is More: Task-aware Layer-wise Distillation for Language Model Compression},
  author={Liang, Chen and Zuo, Simiao and Zhang, Qingru and He, Pengcheng and Chen, Weizhu and Zhao, Tuo},
  booktitle={Proceedings of the 40th International Conference on Machine Learning (ICML)},
  year={2023}
}

@article{DeiT2020,
  title={Training data-efficient image transformers \& distillation through attention},
  author={Touvron, Hugo and Cord, Matthieu and Douze, Matthijs and Massa, Francisco and Sablayrolles, Alexandre and J{\'e}gou, Herv{\'e}},
  journal={arXiv preprint arXiv:2012.12877},
  year={2020}
}

@article{li2016learning,
  title={Learning without Forgetting},
  author={Li, Zhizhong and Hoiem, Derek},
  journal={arXiv preprint arXiv:1606.09282},
  year={2016}
}

@inproceedings{ritter2018online,
  title={Online structured laplace approximations for overcoming catastrophic forgetting},
  author={Ritter, Hippolyt and Botev, Aleksandar and Barber, David},
  booktitle={Advances in Neural Information Processing Systems},
  volume={31},
  year={2018}
}

@inproceedings{schwarz2018progress,
  title={Progress \& compress: A scalable framework for continual learning},
  author={Schwarz, Jonathan and Czarnecki, Wojciech and Luketina, Jelena and Grabska-Barwinska, Agnieszka and Teh, Yee Whye and Pascanu, Razvan and Hadsell, Raia},
  booktitle={International Conference on Machine Learning},
  pages={4528--4537},
  year={2018},
  organization={PMLR}
}

@inproceedings{liu2018rotate,
  title={Rotate your networks: Better weight consolidation and less catastrophic forgetting},
  author={Liu, Xialei and Masana, Marc and Herranz, Luis and Van de Weijer, Joost and Lopez, Antonio M and Bagdanov, Andrew D},
  booktitle={2018 24th International Conference on Pattern Recognition (ICPR)},
  pages={2262--2268},
  year={2018},
  organization={IEEE}
}

@inproceedings{zenke2017SI,
  title={Continual learning through synaptic intelligence},
  author={Zenke, Friedemann and Poole, Ben and Ganguli, Surya},
  booktitle={Proceedings of the 34th International Conference on Machine Learning-Volume 70},
  pages={3987--3995},
  year={2017},
  organization={JMLR. org}
}

@article{yoon2017lifelong,
  title={Lifelong Learning with Dynamically Expandable Networks},
  author={Yoon, Jaehong and Yang, Eunho and Lee, Jeongtae and Hwang, Sung Ju},
  journal={arXiv preprint arXiv:1708.01547},
  year={2017},
  month={August}
}

@inproceedings{jung2020continual,
  title={Continual learning with node-importance based adaptive group sparse regularization},
  author={Jung, Sangwon and Ahn, Hongjoon and Cha, Sungmin and Moon, Taesup},
  booktitle={Advances in Neural Information Processing Systems},
  volume={33},
  pages={3647--3658},
  year={2020}
}

@inproceedings{xu2018reinforced,
  title={Reinforced continual learning},
  author={Xu, Ju and Zhu, Zhanxing},
  booktitle={Advances in Neural Information Processing Systems},
  volume={31},
  year={2018}
}

@inproceedings{rajasegaran2019random,
  title={Random path selection for incremental learning},
  author={Rajasegaran, Jathushan and Hayat, Munawar and Khan, Salman and Khan, Fahad Shahbaz and Shao, Ling},
  booktitle={Advances in Neural Information Processing Systems},
  volume={32},
  pages={},
  year={2019}
}

@inproceedings{gao2022rdfcil,
  title={R-DFCIL: Relation-Guided Representation Learning for Data-Free Class Incremental Learning},
  author={Gao, Qiankun and Zhao, Chen and Ghanem, Bernard and Zhang, Jian},
  booktitle={European Conference on Computer Vision (ECCV)},
  pages={423--439},
  year={2022},
  publisher={Springer}
}

@inproceedings{Choi_2021_CVPR,
  title={Dual-Teacher Class-Incremental Learning With Data-Free Generative Replay},
  author={Choi, Yoojin and El-Khamy, Mostafa and Lee, Jungwon},
  booktitle={Proceedings of the IEEE/CVF Conference on Computer Vision and Pattern Recognition (CVPR) Workshops},
  pages={3543--3552},
  year={2021},
  month={June},
  doi={10.1109/CVPRW53098.2021.00393}
}

@article{hayes2019lifelong,
  title={Lifelong machine learning with deep streaming linear discriminant analysis},
  author={Hayes, Tyler L and Kanan, Christopher},
  journal={arXiv preprint arXiv:1909.01520},
  year={2019}
}

@article{fernando2017pathnet,
  title={Pathnet: Evolution channels gradient descent in super neural networks},
  author={Fernando, Chrisantha and Banarse, Dylan and Blundell, Charles and Zwols, Yori and Ha, David and Rusu, Andrei A and Pritzel, Alexander and Wierstra, Daan},
  journal={arXiv preprint arXiv:1701.08734},
  year={2017}
}

@article{wang2023Survey,
  title={A Comprehensive Survey of Continual Learning: Theory, Method and Application},
  author={Wang, Liyuan and Zhang, Xingxing and Su, Hang and Zhu, Jun},
  journal={arXiv preprint arXiv:2302.00487},
  year={2023}
}

@techreport{cifar100,
  title={Learning Multiple Layers of Features from Tiny Images},
  author={Krizhevsky, Alex},
  year={2009},
  institution={University of Toronto},
  type={Technical Report},
  number={TR-2009}
}

@article{ViT2020image,
  title={An image is worth 16x16 words: Transformers for image recognition at scale},
  author={Dosovitskiy, Alexey and Beyer, Lucas and Kolesnikov, Alexander and Weissenborn, Dirk and Zhai, Xiaohua and Unterthiner, Thomas and Dehghani, Mostafa and Minderer, Matthias and Heigold, Georg and Gelly, Sylvain and Uszkoreit, Jakob and Houlsby, Neil},
  journal={arXiv preprint arXiv:2010.11929},
  year={2020}
}

@article{nguyen2019toward,
  title={Toward understanding catastrophic forgetting in continual learning},
  author={Nguyen, Cuong V and Achille, Alessandro and Lam, Michael and Hassner, Tal and Mahadevan, Vijay and Soatto, Stefano},
  journal={arXiv preprint arXiv:1908.01091},
  year={2019}
}

@InProceedings{Hendrycks_2021_ICCV,
    author = {Hendrycks, Dan and Basart, Steven and Mu, Norman and Kadavath, Saurav and Wang, Frank and Dorundo, Evan and Desai, Rahul and Zhu, Tyler and Parajuli, Samyak and Guo, Mike and Song, Dawn and Steinhardt, Jacob and Gilmer, Justin},
    title = {The Many Faces of Robustness: A Critical Analysis of Out-of-Distribution Generalization},
    booktitle = {Proceedings of the IEEE/CVF International Conference on Computer Vision (ICCV)},
    month = {October},
    year = {2021},
    pages = {8340-8349}
}

@article{kingma2014adam,
  title={Adam: A method for stochastic optimization},
  author={Kingma, Diederik P and Ba, Jimmy},
  journal={arXiv preprint arXiv:1412.6980},
  year={2014}
}

@inproceedings{zhao2022decoupled,
  title={Decoupled Knowledge Distillation},
  author={Zhao, Borui and Cui, Quan and Song, Renjie and Qiu, Yiyu and Liang, Jiajun},
  booktitle={Proceedings of the IEEE/CVF Conference on Computer Vision and Pattern Recognition (CVPR)},
  pages={11953--11962},
  year={2022},
  doi={10.48550/arXiv.2203.08679},
}

@article{romero2014fitnets,
  title={FitNets: Hints for Thin Deep Nets},
  author={Romero, Adriana and Ballas, Nicolas and Ebrahimi Kahou, Samira and Chassang, Antoine and Gatta, Carlo and Bengio, Yoshua},
  journal={arXiv preprint arXiv:1412.6550},
  year={2014},
  eprint={1412.6550},
  archivePrefix={arXiv},
  primaryClass={cs.LG}
}

@inproceedings{chen2021ReviewKD,
  title={Distilling Knowledge via Knowledge Review},
  author={Chen, Pengguang and Liu, Shu and Zhao, Hengshuang and Jia, Jiaya},
  booktitle={Proceedings of the IEEE/CVF Conference on Computer Vision and Pattern Recognition (CVPR)},
  pages={5008--5017},
  year={2021},
  doi={10.48550/arXiv.2104.09044},
}

@InProceedings{APT_2025_ICCV,
    author    = {Chen, Haoran and Wang, Ping and Zhou, Zihan and Zhang, Xu and Wu, Zuxuan and Jiang, Yu-Gang},
    title     = {Achieving More with Less: Additive Prompt Tuning for Rehearsal-Free Class-Incremental Learning},
    booktitle = {Proceedings of the IEEE/CVF International Conference on Computer Vision (ICCV)},
    month     = {October},
    year      = {2025},
    pages     = {340--349}
}

@inproceedings{li2021prefix,
  title     = {Prefix-Tuning: Optimizing Continuous Prompts for Generation},
  author    = {Li, Xiang Lisa and Liang, Percy},
  booktitle = {Proceedings of the 59th Annual Meeting of the Association for Computational Linguistics and the 11th International Joint Conference on Natural Language Processing (Volume 1: Long Papers)},
  pages     = {4582--4597},
  year      = {2021},
  month     = aug,
  address   = {Online},
  publisher = {Association for Computational Linguistics},
  doi       = {10.18653/v1/2021.acl-long.353}
}

@inproceedings{jung2023dap,
  title={Generating Instance-level Prompts for Rehearsal-free Continual Learning},
  author={Jung, Dahuin and Han, Dongyoon and Bang, Jihwan and Song, Hwanjun},
  booktitle={Proceedings of the IEEE/CVF International Conference on Computer Vision (ICCV)},
  year={2023}
}

@inproceedings{gao2024cprompt,
  title={Consistent Prompting for Rehearsal-Free Continual Learning},
  author={Gao, Zhanxin and Cen, Jun and Chang, Xiaobin},
  booktitle={Proceedings of the IEEE/CVF Conference on Computer Vision and Pattern Recognition (CVPR)},
  year={2024}
}

@inproceedings{kim2024osprompt,
  title={One-Stage Prompt-Based Continual Learning},
  author={Kim, Youngeun and Li, Yuhang and Panda, Priyadarshini},
  booktitle={Proceedings of the European Conference on Computer Vision (ECCV)},
  pages={163--179},
  year={2024}
}

@techreport{wah2011caltech,
  title       = {The Caltech-UCSD Birds-200-2011 Dataset},
  author      = {Wah, Catherine and Branson, Steve and Welinder, Peter
                 and Perona, Pietro and Belongie, Serge},
  institution = {California Institute of Technology},
  number      = {CNS-TR-2011-001},
  year        = {2011}
}
